# Prosody: Rhythms and Melodies of Speech

Dafydd Gibbon

Bielefeld University, Germany

**Abstract**

The present contribution is a tutorial on selected aspects of prosody, the rhythms and melodies of speech, based on a course of the same name at the Summer School on *Contemporary Phonetics and Phonology* at Tongji University, Shanghai, China in July 2016. The tutorial is not intended as an introduction to experimental methodology or as an overview of the literature on the topic, but as an outline of observationally accessible aspects of fundamental frequency and timing patterns with the aid of computational visualisation, situated in a semiotic framework of sign ranks and interpretations. After an informal introduction to the basic concepts of prosody in the introduction and a discussion of the place of prosody in the architecture of language, a stylisation method for visualising aspects of prosody is introduced. Relevant acoustic phonetic topics in phonemic tone and accent prosody, word prosody, phrasal prosody and discourse prosody are then outlined, and a selection of specific cases taken from a number of typologically different languages is discussed: Anyi/Agni (Niger-Congo>Kwa, Ivory Coast), English, Kuki-Thadou (Sino-Tibetan, North-East India and Myanmar), Mandarin Chinese, Tem (Niger-Congo>Gur, Togo) and Farsi. The main focus is on fundamental frequency patterns. Issues of timing and rhythm are discussed in the pre-final section. In the concluding section, further reading and possible future research directions, for example for the study of prosody in the evolution of speech, are outlined.

## 1    Rhythms and melodies

Rhythms are sequences of alternating values of some feature or features of speech (such as the intensity, duration or melody of syllables, words or phrases) at approximately equal time intervals, which play a role in the aesthetics and rhetoric of speech, and differ somewhat from one language or language variety to another under the influence of syllable, word, phrase, sentence, text and discourse structure.

    Melodies are contours of the pitch values associated with syllables, words and whole utterances, and contribute to rhythms whenever their pitch patterns alternate in similar time intervals, but also have additional properties of rising, falling or level pitch with their own functionalities.

    Rhythms and melodies which contribute to language structure and meaning constitute the domain of *prosody*. Related phonetic properties of voice quality which mark emotional or physical states and individual speaker characteristics are often known as *paralinguistic features*, but the boundaries between paralinguistic features and the prosodic features which contribute to language structure and meaning are somewhat fluid. The term 'paralinguistic' is also sometimes used to refer to non-verbal, gestural communication, where analogies to the rhythms and melodies of speech are found.

    There are many motivations for studying prosody and many disciplines in which the study of prosody has proved to be essential. The most conspicuous disciplines in this respect are: linguistics (in relation to language structure and form); phonetics (in the causal sequence from the *phonation rate* of the larynx in speech production through the *fundamental frequency*, *F0*, of the speech waves in transmission to the impression of *pitch* in perception); the philosophy of language (in relation to speech acts); sociolinguistics and sociology (in relation to interactive social behaviour); psycholinguistics (in relation to speech behaviour and the cognitive processing of speech); gestural studies (in relation to the related functionalities and parallel structures of gesture and prosody); clinical linguistics, clinical phonetics and speech therapy (in diagnosis and therapy of speech disabilities resulting from a range of conditions from strokes to Parkinson's disease); language learning and teaching (in relation to the different prosodic structures of source and target languages); evolutionary linguistics (in the relation of speech prosody to the rhythms and melodies of the calls, cries and hoots of primates and other species); speech technology (in relation to human-machine interaction).

    This tutorial on prosody is based on course materials used at the *Summer School on Contemporary Phonetics and Phonology* at Tongji University, Shanghai, China, in July 2016. The tutorial is aimed at intermediate level linguistics

students and other readers who have a serious interest in learning about the rhythms and melodies of speech. Some aspects of the modelling of speech melody and prosodic functionality may also be of interest to the specialist. The tutorial is not intended as a survey of the literature on the topic, though guidelines to relevant literature are provided, especially in the concluding section. Nor is the tutorial intended as an introduction to experimental methodology, though some relevant methods are illustrated. The main objective is to discuss applications of basic computationally supported acoustic phonetic techniques for empirical observation of core *semiotically relevant* aspects of prosody at different *hierarchical ranks* in the *architecture of language*. The ranks are hierarchically arranged systems of units of different sizes in language and speech, each with its own unique structure, from discourse (including dialogue) through texts, sentences, phrases and words to phonemes. The term 'rank' has been used previously in linguistics for similar concepts, but the reason for selecting this term in the present context is that *constraints at higher ranks may override constraints at lower ranks*. The term *rank* in the context of the architecture of language has a different meaning from *rank* or *ranking* in Optimality Theory, where it means the order of application of constraints on surface realisations of underlying structures (cf. Kager 1999).

The theory of the phonological hierarchy formulated by Selkirk (1984) and its later variants is in general relevant to F0 patterning, but is not treated in this tutorial, where the focus is on the phonetic grounding of prosody, particularly of tone, accent and intonation, than on phonology. Detailed phonetic properties of the phonological hierarchy through all ranks in the architecture of language have yet to be worked out.

Most attention is paid in the tutorial to melody, specifically to the phonetic trio of articulation rate in speech production, fundamental frequency (F0) in speech transmission, and pitch in speech perception, as well as to the functionality of melody. There are two main reasons for this restriction. One reason is practical: there are easy to use computational tools for analysing speech, including F0, which are available for download on the internet as freeware and open source software. A second reason for concentrating on melody is that, in addition to its contribution to rhythm, melody has many other functions at different ranks in the architecture of language, while rhythm turns out on careful and detailed analysis to be essentially a discourse issue. Sometimes the 'rhythm of words' and the 'rhythm of phrases' is discussed in the literature, but these terms are better understood as referring simply to the phonological and morphological structure of words and phrases and its role in speech timing in general, not necessarily to rhythm in the usual sense of alternations of similar events at regular intervals.

Rhythm in this strict sense is difficult to pinpoint in the physiological and physical domains. To a large extent speech rhythm appears to be a complex temporal organising principle operating in the cognitive processing of percepts: rhythms may even be superimposed on purely mechanical event repetitions. Discourse rhythms and discourse melodies may override structurally determined lexical and phrasal timing factors such as simple or complex syllable structure, simple or complex morphological structure, relations between functional and lexical words in phrases, and dependency relations in sentences. Similarly, the well-defined metres of traditional poetry and the time signatures of music may be overridden by rhetorical and aesthetic syncopation in performance. Nevertheless, there are clear cases where speech rhythm is indeed physically grounded in measurable timing patterns. Some of these physical properties of speech timing are discussed towards the end of this tutorial paper.

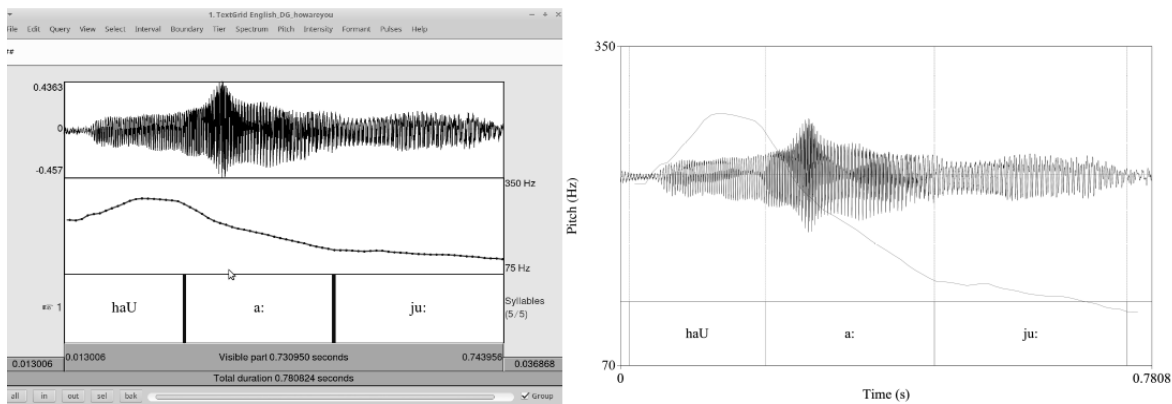

Figure 1: Example of Praat visualisations with screenshot of graphical signal editor interface (left) and drawing output (right) of waveform, F0 analysis and transcription labels.

Visualisation of acoustic properties of speech by means of automatic annotation mining (Gibbon and Fernandes 2005) is emphasised in this context, on the principle that (at least in a tutorial) 'one picture is worth a thousand words'. The most well-known and widely used software tool for visualisation and analysis phonetics, including prosody, is *Praat* (Boersma 2001), which is a very comprehensive workbench for acoustic phonetic analysis, visualisation and experimentation, with its own programming language, 'Praat scripting', which lends itself to automatic annotation mining as well as manual analysis. In Figure 1 an example of the Praat user interface (left) and the picture drawing facility (right) is shown with a recording of an utterance of "How are you?", with the waveform (top), the F0 trace (centre), the annotation labels or TextGrid, the time, frequency and amplitude parameters and window handling information. Generally the picture drawing is recommended for publications, but many authors prefer the greater detail of screenshots. The annotation file (known as a TextGrid) can be saved in a text format with labels and time-stamps relating the transcription to times in the speech recording. The annotation data in text format can be mined for statistical and structural information, using spreadsheet software or a specially written software application.

There are many other useful and easy to find software tools such as *SPPAS* (Bigi 2015) for semi-automatic annotation, *Annotation Pro* for annotation and experimentation (Klessa 2016), as well as general purpose audio editors like *Audacity* and the powerful command line audio tool *sox* (Poser 2015). There are also online tools such as *TGA* for annotation mining of timing patterns in annotated speech recordings (Gibbon and Yu 2015).

The *Prosody Visualisation* (*PV*) software which has been used for most of the figures in the present tutorial employs annotation mining of Praat TextGrid files to support analysis of the speech signal as well as analysis of the annotation itself. *PV* was designed specifically for use in the Tongji University Summer School and for use in this tutorial in order to provide features which are not easy to obtain with other tools. These features include (1) display of F0 which is much longer than the 10 seconds of Praat default visualisation; (2) implementation of models of F0 visualisation, including *polynomial stylisation*; (3) visualisation on any graphical browser on any computing platform via a web interface. The *PV* tool is not an annotation tool but a specialised visualisation tool only, and is not intended to replace other tools. The *PV* tool will be made available as soon as possible for general use.

Some information in the tutorial is repeated in different sections, not a bad thing in a tutorial. As a guide for readers, salient terms, including but not restricted to technical terms, are rendered in italics. Depending on the context, examples from relevant languages are rendered either in quotation marks or italics.

The tutorial is structured as follows. After the brief overview in this introductory section, the second section discusses basic properties of prosody with particular reference to terminology and basic phonetic information. The following sections deal with modelling prosody in the acoustic phonetic domain at different ranks in the architecture of language, starting with lexical (phonemic and morphological) prosody and moving through phrasal prosody to discourse prosody. The final section provides a summary, a conclusion and an outlook on possible future developments, including the role of prosody in the evolution of speech.

## 2  Speech melodies: structure and function

### 2.1  Prosodic terminology: 'tone', 'pitch accent', 'stress', 'intonation'

In the present contribution, the terms 'melody' and 'rhythm' are used as informal, intuitively clear terms, functioning as *explicanda* which are neutral with respect to the technical definitions and explanations provided later. The technical definitions are intended to be a heuristic guide in the often rather confusing terminological wilderness encountered in the literature on prosody, and may thus appear too simple to the prosody specialist.

The term 'suprasegmental' is often used as a partial synonym of 'prosody', setting the relevant features in relation to 'segmental' units, i.e. phonemes, and implying that the features concerned are 'above' or 'larger than' phonemes. Except for phonemic tones and phonemic pitch accents, prosodic forms are not in great need of phoneme segments as a reference point, however, and there are independent criteria for describing and modelling them. Indeed, utterances which consist of single phonemes have prosodic properties of timing and melody, so temporal size is neither a necessary nor a sufficient criterion: "Ah!", "Oh?", "Mmm...". A more congenial alternative term is 'autosegmental' (Goldsmith 1990), referring to the parallel structuring of locutionary items such as syllables, words, phrases, sentences, and prosodic items. But in this tutorial the term 'prosody' is used.

The terms *tone* and *pitch accent* are used ambiguously in the literature. In the first instance they apply to the assignment of a specific melody to a word to make a *phonemic distinction* or add a *morphological function* and are thus associated with the domain of *lexical prosody*. The basic difference between tones and pitch accents is that *a tone is one of a set* of melodies, as in Mandarin Chinese, while *a pitch accent is basically a single melody* which may be assigned to different lexically distinctive positions in a word (as in Japanese, cf. Poser 1984). The differences are not clear-cut, however, as Hyman (2009) has pointed out, and details differ greatly from language to language.

The term *pitch accent* is also often used with a different meaning, for a set of melodies which may be assigned to different lexically distinctive *stress positions* in a word (as in English), but whose melodies have semantic and pragmatic rather than lexical functions, indicating old and new information, or contrast and emphasis. In addition to their association with stress positions in words and phrases, they also function in larger contexts where they merge with utterance-initial or utterance-final boundary tones and with overall utterance-length pitch patterns, which also have pragmatic functions relating to speech acts and to dialogue management patterns such as turn-taking.

The English type of pitch accent is associated with *stress positions* in words and phrases, which are a more abstract feature of language structure than their phonetic realisations as pitch accents. The stress positions in words (*word stress*) or in phrases and sentences (*phrase stress*, *sentence stress*, *nuclear stress*) are the structural locations to which pitch accents are assigned. The pitch accents themselves have semantic and pragmatic functions. Confusingly, the term *accent* is also used in the literature with this abstract positional meaning, e.g. as *word accent*, *sentence accent*. The term *stress* is frequently used in the literature with a phonetic rather than structural meaning, along with terms such as *prominence* and *salience*. The definitions used here are designed to provide a coherent overall terminological framework for an ontology of prosody, which is lacking in the literature (cf. also Jassem and Gibbon 1980; Gibbon 1995; Gibbon 2006).

The term *intonation* refers to rhythms and melodies which are assigned to language constituents which are larger than the word, such as the global contour of a phrase, sentence, utterance or utterance sequence in discourse, as well as the melodic patterns of English-type pitch accent and boundary tone sequences, the higher, lower, falling and rising melodic patterns and the accelerating or decelerating rhythmic patterns of whole utterances. The term *tone* is also found in the literature with the non-lexical meaning of pitch accent at constituent ranks above the word, as in the *ToBI* (*Tone and Break Index*) transcription system (Beckmann et al. 2005), or in the terminologically elegant systemic intonation model of Halliday: *tonality* (syntagmatic domains of intonation), *tonicity* (placement of accents in phrases or sentences) and *tone* (the final or pre-pausal pitch accent in a phrase, sentence or utterance); cf. Halliday (1967). Patterns which merge more than one intonation unit into a two-level 'flat hierarchy' have been variously referred to as *paragraph intonation units* (Lehiste 1975; Wichmann 2000) or *paratone* units (Brazil et al. 1980).

Because of the multi-rank functionality of tone and pitch accent at the lexical rank, and of intonation units and paratones at higher ranks, an interesting problem arises if an utterance is rather short, perhaps consisting only of one word, such as [*Where is it?*] *There!* In this case, higher ranking intonation patterns may modify or override the tone or pitch accent, making it difficult to separate word, phrase, sentence and utterance prosody from each other. This complex constellation arises most clearly in citation contexts, for example, when a *faux* dialogue context is created for the purpose of illustration or linguistic argument. A sentence like *Willy married Mary*, for example, may be spoken as neutral information or with contrastive or emphatic accents in many kinds of dialogue citation context: "[*After a long engagement,*] *Willy married Mary*", "[*JACK didn't marry Mary,*] *WILLY married Mary*", and many other stress positioning options. A citation context as used in these examples is an idealised context which is constructed specifically for the purpose of analysis or illustration of a citation form such as the capitalised forms in these examples. Citation contexts contrast with more complex authentic or realistic speech production contexts. Dialogue citation contexts are sometimes referred to in the present tutorial as *faux* (false) dialogues.

There are many kinds of notation which are used to represent stress positions, pitch accents, tones and intonations, ranging from upper case letters for stress or accent, as in *WILLY married MARY*, to accent and tone diacritics and numbers (cf. Pinyin transcriptions of Mandarin Chinese tone 1 as *fēng* or *feng1*, wind (just one of several meanings). In English, lexical stress position is commonly represented as *imPErative* or, with a diacritic, *impérative*, or, in IPA phonemic transcription, /ɪmˈpɛrətɪv/. The International Phonetic Alphabet offers a basic range of symbols for prosody, Hirst's *IntSInt* (Hirst and Di Cristo 1998) is used in computational phonetic modelling, and the *ToBI* system is widely used for phonological modelling both in linguistics and in speech annotation for technological applications. In the applied phonetic teaching literature there are 'icon' notations in which the symbols used for intonation and tone are intended to represent the perceived pitch movements, ranging from text type-set in wavy lines to graphics with dots, and with tails on the dots indicating pitch direction ('tadpole notation'). A fundamental issue underlying the choice of such symbol systems is not only the purpose of the transcription but also the ontology of intonation and tone, i.e. the choice of basic entities in terms of which intonation and tone are described. One of the most basic distinctions, contrasting 'pitch level' models with 'pitch contour' models (cf. the uses of 'H-L' or '2-1' with similar meaning to '\') was first debated in detail by Bolinger (1951).

## 2.2 A note on 'prosody'

The term *prosody* derives from the ancient Greek word προσῳδίᾱ, /prɔsoːˈdiaː/, which had two meanings: either syllabic accent, or song. These meanings set the scene for the contemporary meanings of *prosody* in linguistics and poetics. The meaning of *poetic prosody* in the study of literature refers to genre-specific patterns of *metre* in poems, for example regularly occurring patterns of prominent (e.g. strong, accented or long) syllables and less prominent (e.g. weak, unaccented or short) syllables. Poetic metres define basic rhythm types, which may be varied and combined or overridden for aesthetic and rhetorical effect in performance. The basic two-part unit of rhythm, the *foot*, with a weak-strong syllable sequence (commonly annotated ⌣– or ×/) is termed the *iamb* or the *iambic foot*, while a foot with a strong-weak pattern (commonly annotated –⌣ or /×) is called the *trochee*, or the *trochaic foot*. The phonetic details are different from language to language: in English the 'strong-weak' distinction is determined by lexical or phrasal stress position and pitch accent, while in Latin syllable quantity, i.e. essentially duration, was the criterion. The following quotations show typical examples of traditional poetic metre in English.

> Iambic: "If MUsic BE the FOOD of LOVE, play ON" (W. Shakespeare, *Romeo and Juliet*)

> Trochaic: "DOWN the LANE came MEN in PITboots" (Philip Larkin, *The Explosion*)

As already noted, in actual performances the stress positions indicated by the metre may be overridden by other rhetorical and aesthetic discourse patterns. Alternative stress positions might be, for example: "If MUsic be the food of love, play ON", or "If MUsic be the food of LOVE, play ON", but presumably not "If music be the FOOD of love, play ON".

There are many more named metrical patterns (28 were in the specialist literature), the most common of which, in addition to the *iamb* and the *trochee* are the *anapaest* ⌣⌣– (*it's a boy*), the dactyl –⌣⌣ (*happily*), the spondee – – (*Good boy!*) and the amphibrach ⌣–⌣ (*Carola*). The terms *iambic*, *trochaic* and *foot* have found their way into linguistics and are accepted metaphorical technical terms in phonology, with *iambic* and *trochaic* applying to weak-strong and strong-weak feet, respectively.

In linguistics, perhaps influenced by these metaphorical terms, timing patterns are often taken to be binary, either iambic or trochaic. This may be a common default case, but the realities of language structure and speech production are different, as shown by patterns in English like *JONathan | APPleby | CAME on a | BIcycle* (dactylic), *It's a SHAME | that he CAME | on his OWN* (anapaestic), *my EYES danced | a CIRcle | aCROSS her | clear OUTline* (amphibrachic, from *Eternal Circle*, by Bob Dylan), or even monomacric, i.e. with 'syllable timing': ADD | DEATH | WHAT | SHALL (e.e. cummings, *Being*). Expressions are sometimes ambiguous in terms of these types, and assignment to these structural types may therefore vary according to whether lexical, morphological or phrasal criteria, or rhetorical and aesthetic rhythm dominate segmentation into feet. The stress position patterns, rhythms and melodies of speech are not constrained by such pre-defined pattern inventories: the field of *speech prosody* is much broader, covering several flexibly structured ranks of timing and melody patterning, ranging from phonemic tonal distinctions to the rhythms and melodies of discourse.

The present contribution is motivated by an interest in the physical grounding of the pattern production and recognition capabilities used in human communication, on the one hand, and by the semiotic function of prosody, on the other. Many of the concepts introduced in this section are fairly common knowledge in linguistics and therefore do not need copious citation references.

Sometimes it is claimed that prosody is not as well researched as other areas of language and speech. But this is not true. A quick look at the literature on prosody reveals a very broad spectrum of publications in a very wide range of scenarios. These scenarios range from textbooks on topics such as 'stress' and 'intonation' in foreign language teaching and on the 'tone of voice' in rhetoric primers for native speakers, through linguistic treatments of 'intonation', 'tone', 'accent' and 'stress' in relation to discourse, sentence and word structure, and to experimental studies of timing and frequency trajectories in phonetics, pyscholinguistics and speech engineering. The literature on speech prosody is in fact enormous. But terminology has developed rapidly and independently in different places, and therefore there is a sometimes rather confusing range of not-quite-similar terms for not-quite-similar concepts.

The peer-reviewed conference series *International Congress of Phonetic Sciences* and *Speech Prosody* are contributing to the stabilisation of terminology, but terms such as *tone*, *boundary tone*, *pitch accent*, *accent*, *stress*, *prominence*, *salience*, *focus*, *sentence stress*, *nuclear stress*, *word stress*, *intoneme, intonation morpheme*, *intonation group*, *breath group*, *sense group*, *thought group* occur in the literature with a profusion of overlapping, or different, or conflicting meanings. To provide a useful frame of reference for the present contribution, a selective systematic terminology is used, based on a model of ranks in the architecture of language, each of which is structured into three main dimensions of speech signs.

## 2.3 Melody in the architecture of language

### 2.3.1 Prosodic signs

The core of the semiotic model which forms the foundation of descriptions and explanations in this tutorial is shown in Figure 2. The model has three dimensions:

1. *Structure*: the linguistic inventories, categories and relations, sequences, parallel structures and hierarchies which organise language (whether written, spoken or signed) into recognisable patterns. Prosody in the structural domain is referred with the terms *lexical tone*, *morphological tone* and *intonation*, in terms of *stress positions*, in particular *lexical stress positions*, *phrase stress positions* and *sentence stress positions*, but also in terms of stress positions determined by the structures of text and discourse. Stress positions are associated with the *prominence* or *salience* provided by *pitch accents*, i.e. local melodies, and by *quantity*, i.e. by variations in syllable and word duration, and to some extent by the relative *intensity* of the speech signal.

2. *Modality* (form, substance): the physical transmission path for speech, from the neurophysiology of the speaker via the speech production organs and the transmission medium of air and intervening electronic channels to the speech reception organs and the neurophysiology of the listener. The concept of modality is multidimensional: visual modalities are used for writing and signing, and combinations of these are used in the multimodal communication of speech and in conversational gesture, and of speech which in some contexts accompanies writing. The term *multimodal* communication refers to communication which combines phonetic, visual and tactile modalities; the term *multimedia* communication refers to the use of different technical media in the transmission phase of communication. Melody in the phonetic modality domain is referred to in terms of *phonation rate* in the phonation subdomain, *fundamental frequency* (*F0, F zero*) in the transmission domain, and *pitch* in the perceptual domain. Prosody is somewhat independent of the locutionary components of speech, and prosody and locutions may be thought of as parallel, simultaneous *submodalities* of speech.

3. *Function*: the semiotic value of signs, including *semantic meaning* involving objects, events, facts and properties, with semantic concepts such as *information structure* or *focus* and *background*, and *contrast* being particularly relevant to prosody, and *pragmatic meaning* involving personal features and interpersonal relations, with *emphasis*, *attitudes*, *feelings*, *emotions*, *sentiments*, *empathy*, *irony* and *sarcasm* being particularly relevant to prosody.

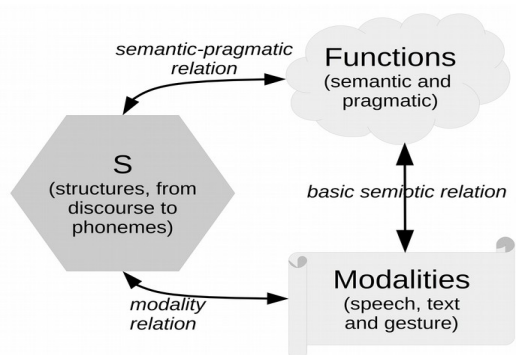

Figure 2: Three component model of relations between sign structures, functions and modalities.

The domains of Modality, Structure and Function apply to all ranks of communicative units, from the rank of phoneme through the morpheme and the word, and through phrases, sentences and texts, to the rank of interpersonal discourse. Some models of language and speech only have two components, corresponding to Modality and Function. Other models are often restricted to words and sentences and to either written or spoken language, and do not reach as far as the discourse domain. Some models are also tripartite, like the present semiotic model, for example the Chomskyan model of syntax with phonetic interpretation (a subdomain of the semiotic Modality domain) and semantic interpretation (a subdomain of the semiotic Function domain), but are highly constrained to cover only word and sentence ranks. These semiotically restricted models are inadequate for handling the range of prosodic structures and meanings, and indeed for handling the full range of locutionary structures and meanings, too. For a reasonably full discussion of prosody, all three semiotic dimensions as well as the whole range of ranks from discourse to phoneme are needed.

In summary, both *rhythm* and *melody* involve sequential and hierarchical patterning of syllables, words and phrases in time, and interact in various ways, both with each other and with the expressions they accompany. Both rhythms and melodies are more than patterns of sound: they are *signs* with their own specific *semiotic value* in terms of pragmatic, semantic and structural meanings. The meanings of the rhythms and melodies of speech partly overlap with the meanings of locutions, i.e. words and phrases. In different ways and in different languages the rhythms and melodies of speech help to create interactive speech acts and speech act sequences, for example of announcing, questioning, answering or commanding. Speech styles such as informal conversation or formal styles such as political speeches are marked by different rhythms and melodies, and rhythms and melodies contribute to the expression of attitudes such as cheerfulness, scorn, anger or uncertainty. These functions can, of course, also be expressed without taking rhythms and melodies or conversational gestures into account, with the appropriate choice of vocabulary, verb forms and intensifying or deprecating terms. However, locutionary meanings can be contradicted in specific situations by prosody, for instance in *ironic* and *double bind communication*, in which a positive impression is intended to be communicated in a disapproving context.

In conjunction with background information from the communication context, rhythms and melodies may contribute to the disambiguation of ambiguous locutions. In addition to the concomitant, supporting semiotic functions of prosody, rhythms and melodies also have their own specific *metalocutionary* semiotic properties. The metalocutionary properties include intensification of the locutionary meanings which they accompany, but may also be somewhat independent. Pitch accents may accompany demonstrative pronouns and adverbs with *deictic* or pointing functions, as in *I meant THIS one over HERE, not THAT one over THERE*. But in their own *metadeictic* functions rhythms and melodies denote – in the same strict semantic sense of denotation – temporal locations of locutionary constituents in utterances: the temporal boundaries of phrases, sentences and utterance turns, the temporal locations of words and phrases with new and old information, indicating contrast and emphasis. For example, the pitch accents associated with stress positions in *DAFfodils are her FAvourite FLOWers*, have the semantic role of pointing to the syntagmatic positions of the nouns *daffodils* and *flowers*, and the adjective *favourite* in the sentence, in addition to other semantic and pragmatic roles they may have.

### 2.3.2 Prosodic ranks

Prosodic signs are organised at the ranks of discourse, text, sentence, phrase and word, with the latter further structured into the 'duality' of morphological structures and functionalities, relating to inflection and word formation, and phonemic structures and functionalities, relating to basic syllable and word structure.

Typical cases of lexical prosody, in which prosodic features of melody and timing are part of word structure, are the Mandarin Chinese lexical tones (*mā*, mother; *má*, hemp; *mǎ*, horse; *mà*, scold; *ma*, question particle), and lexical stress position distinctions in English such as the noun *IMport*, meaning something imported and the verb *imPORT*, meaning the action of importing. Another English example is the pair *INsight* (trochaic) and *inCITE* (iambic). Stress positions are obligatory and partly unpredictable lexical features of English, but cases of actual phonemic distinctiveness of stress positions, and thus also of the phonemically distinctive pitch accents associated with them, are the exception rather than the rule, and are rather rare in the English vocabulary, unlike tone in Mandarin Chinese.

Morphological and phrasal prosody marks prominent syllables, word constituents, words and phrases and phrase boundaries as well as completion or non-completion of constructions such as lists. Clear examples are provided by compound nouns and noun phrases in English: *Chinese geography expert* may mean an expert on Chinese geography, where *Chinese geography* is a nominal expression containing an adjective modifying a noun, and is part of a compound word. Or it may mean an expert on geography who is Chinese, where *geography expert* is the compound noun and the adjective *Chinese* modifies the compound noun. An example of disambiguation by phrasal prosody may occur with ambiguous locutions like *old men and women*, which may refer to old men and to all women, or to old men and old women. Such distinctions may be communicated by indicating structural boundaries with slower timing, for example by syllable lengthening and pausing at phrase boundaries, as well as by the use of melodic prominence. So in the second example, 'old' may be lengthened and 'men and women' may be spoken more rapidly; in the first example, on the other hand, 'men' may be lengthened or separated from 'women' by a short pause. Like other locution-prosody relations, prosodic disambiguation is optional and may be overridden by discourse factors.

Repeating and expanding some of the points made previously, examples of phrasal prosody occur in lists and other conjunctive sequences, where the melody may indicate noncompletion or completion of a locution in addition to realising a nuclear pitch accent: *let's get some /FISH, /PEAS and po\TAtoes* (where '/' indicates rising melody and '\' falling melody). In the case of */FIRST we go to /ROME and /THEN we go to \PAris* the rising melody may have a second delimitative function in addition to the pitch accents: on *first* and on *then* it indicates the *topics* of the two conjuncts, while on *Rome* it indicates the end of an incomplete locution, and a falling melody on *Paris* may indicate completion of the locution. These are not hard-and-fast rules: they are choices which may be made, and which may be overridden by other higher ranking semantic and pragmatic constraints.

At the discourse level, semantic factors may also override lexical and phrasal factors in stress position assignment. One semantic factor is contrastive meaning, as in *first \WE (not you) go to Rome and \YOU (not we) go to Paris*, may override a straightforward informative interpretation. A pragmatic factor, on the other hand, is emphasis, indicating a speaker attitude or evaluation, perhaps accompanied by other markers such as interjections or intensifying adverbs: *WOW! First we go to \ROME and then we even go to \PAris'*. Discourse prosody may mark pragmatic events such as speech acts and adjacency pairs such as question-answer relations of different kinds, and also discourse control events such as turn-yielding and turn-requesting, or social relations such as polite deference or subordination. Structures consisting of locutions ending in a rise followed by locutions ending in a fall (including question-answer relations and the 'deferential rise'), may be interpreted as a turn-requesting noncompletion-completion sequence, or combine this with a community convention, as in 'uptalk'. In a given case is often hard to disentangle the many potential meanings associated with different ranks.

#### 2.3.3 The neighbours of prosody

So far, selected areas of prosodic form and function have been outlined. There are many other features of pronunciation which interact with prosody and which may be mentioned in passing but not dealt with in detail. For example, when the melody falls to a very low level, the mode of vocal production may turn into *creak* or *creaky voice*, also known as *vocal fry*, in which the vocal folds vibrate more irregularly than in regular sound production. Creaky voice may frequently be observed in both male and female voices in Mandarin Chinese tone 3, the falling-rising tone, in which creaky voice very often occurs at the lowest point of the tone. Creaky voice may also occur at the end of utterances with a low falling melody, where the lowest frequencies may be replaced by creaky voice. More extensive use of creaky voice as a sociolinguistic marker can be observed in many contexts. The creaky voice variant of low F0 features in some of the following sections.

Prosodic functionality may also be observed with other qualitative features of pronunciation such as the lip-rounding which may occur in baby talk with babies and small children. Functionality of this kind is usually described as paralinguistic rather than prosodic.

The term 'prosody' is also sometimes used metaphorically for other communication modalities. Punctuation, highlighting and layout may be seen, from a functional semiotic point of view, as the prosody of written text. In conversational gesture and sign language communication, aspects of the metaphorically termed 'sign language phonology' which indicate semantic content, such as deictic pointing, and pragmatic attitudes and emotions, are sometimes referred to by the metaphor of 'gesture prosody'.

In *Prosodic Analysis*, the approach to phonology and prosody of Firth and the London school (Palmer 1968, 1969), the term 'prosody' refers to syntagmatic properties of speech sound sequences which cover more than one phoneme and are parallel to sequences of *phonematic units*. For example certain *allophone contexts*, *assimilation between neighbouring sounds*, *vowel harmony*, *tone* and *tone sandhi* as well as stress positions, intonation and pitch accents, count as prosody. An approach which is somewhat related to the Firthian approach in analysing many syntagmatic features of phonology as occurring in parallel with to a 'segmental' or phonemic string is *Autosegmental Phonology* (Goldsmith 1990). An approach which concentrates on the structures and roles of stress position hierarchies is *Metrical Phonology* (Liberman 1975; Hayes 1981). Both Autosegmental and Metrical phonologies figure extensively in the literature, and Ladd (2009) combined the approaches into *Intonational Phonology*.

## 3 Melody in the phonetic modality

### 3.1 The speech cycle

Physically, speech is a complex causal process with five main phases: speech planning, speech production, speech transmission, speech perception and speech comprehension. Each phase implements a transformation of the speech signal into the next phase, and also involves motor and sensory subphases which are not dealt with in detail here. Each phase is associated with specific terminology concerning prosodic parameters, some of which was noted in the previous section: *effort*, *phonation rate* in producing the fundamental frequency and *articulation rate* of speech sounds (speech production); *amplitude*, *intensity*, *energy* and *fundamental frequency* (speech transmission); *loudness* and *pitch* (speech perception). These contrast with linguistic terms such as *stress position*, *accent*, *tone*, *intonation*, *paratone* for structural categories, and *focus*, *contrast*, *emphasis*, *speech act marker* for semantic and pragmatic interpretations. Details of speech production, transmission and perception which relate more to phonemic speech sounds and their features than to prosodic features are not described in this tutorial.

The planning and comprehension phases include both the planning of rhythms (timed by the 'brain clock' and its embodiment in physiological timing properties of production and perception) and the planning of melodies. Planning and comprehension are cognitive processes which take place in the brain. The speech production and perception phases are anchored in the planning and comprehension processes in the brain and extend along efferent and afferent nerves to the domain of articulatory tract physiology in the throat and mouth and the auditory physiology of the ear. The cyclical relation between the five phases – in which of course other communication partners may intervene – is illustrated in Figure 3.

Speech processes in the brain are only partly understood, and derive mainly from investigations of speech disturbances, leading to results such as the assignment of pragmatic aspects of prosody to the right hemisphere of the brain and structural aspects to the left, the discovery of differences in prosody in Wernicke and Broca aphasias, as well as characteristic prosodic features of Parkinson's disease and other neurological illnesses. Models of the brain with properties which are directly relevant to properties of prosody and other properties of speech are relatively coarse-grained in comparison with models of the domains of speech production, transmission and reception.

The speech production process is quite well understood. Models of articulation processes and their acoustic properties have been developed in *articulatory phonetics*, and high quality artificial speech synthesis systems have been based on these models. Speech perception in *auditory phonetics* is also very well understood in terms of the transmission of speech signals in four phases within the ear: directional signal reception by the *auricle* or outer ear, transmission via the *ear canal* and the *eardrum* to the middle ear, where pressure transformation by the *ossicles* ('little bones') suspended in the inner ear takes place, with further transmission via the *oval window* to the *cochlea* ('snail') or inner ear, where the frequencies of speech are transformed into nerve signals and transmitted to the brain.

The term *auditory phonetics* refers to the study of the ear and its functionality, but it also has a second meaning, referring to the method of analysing speech signals by means of listening and describing the volume, 'brightness' and 'darkness', and 'tempo' of what is heard. Auditory phonetics in this sense is often contrasted with instrumental phonetics, in which measuring instruments as well as computational analysis and visualisation are used, and with experimental phonetics. The contrast is not a true contrast, because instrumental and experimental methodologies in phonetics presuppose basic prior understanding of auditory phonetic properties before analysis with measuring instruments and computational analysis. In experimental phonetics perceptual judgments of properties of speech signals are an important object of research.

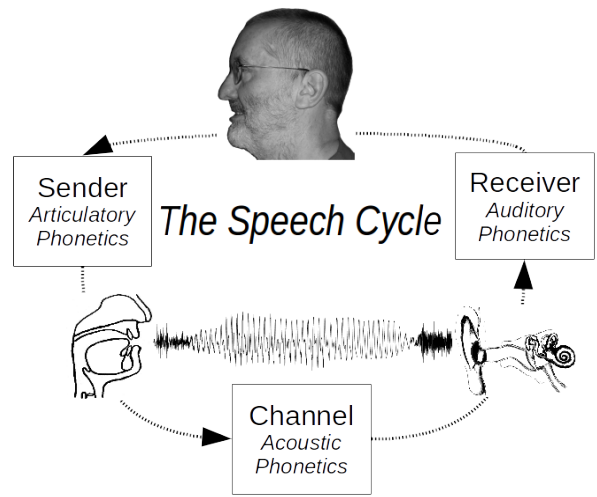

Figure 3: The basic speech cycle. The channel includes communication with interlocutors and also technical media.

In speech production the correlate of melody is phonation rate, the frequency of vibration of the vocal folds in the larynx in the *modal voice phonation* type. There are many easily accessible and reliable sources of information about the anatomy and physiology of the larynx and this information is not duplicated here. A brief but detailed and helpful overview is given by Fitch and Hauser (1995) in their discussion of prosody in nonhuman and human primates. Other phonation types are: *unvoiced* (vocal folds separated and not vibrating); *creaky voice* at low fundamental frequencies (for apparently unknown reasons also known as *vocal fry*; speculations abound that it may be a metaphor referring to the sound of frying fish and chips). Other types of phonation are *whisper* (practically no vibration of the vocal folds), *breathy voice* (sometimes seen as a combination of modal voice and whisper), *harsh voice* (characterised by higher tension in the larynx) and *falsetto* (used in yodellling, for example: the vocal folds are stretched beyond their normal range and do not close completely, resulting in a frequency which is about an octave higher, somewhat like a guitar or violin string when lightly touched exactly in the middle).

In the phonation types unvoiced and whisper the vowel formants are preserved in the noise created by air friction in the larynx or from consonants formed higher in the vocal tract, but the fundamental frequency is absent. In creaky voice phonation the regular low fundamental frequency line is destroyed, and replaced by an even lower, usually irregular frequency; the irregular vibrations can also be detected as irregularities in the waveform of the speech signal.

Phonation rate and phonation type are controlled by four main parameters: distance between the vocal folds; stretching the length of the vocal folds; changing the stiffness and thickness of the vocal folds; raising and lowering the larynx. The raising and lowering of the larynx can be detected to some extent externally, by placing the fingers on the front of the neck and imitating a siren with lowering and raising of the melody.

The type of vibration of the vocal folds which has the greatest relevance for most studies of prosody is the modal voice type, with the vocal folds producing a regular type of vibration with a fundamental frequency and a lengthy series of *harmonics*, which are higher frequencies which are integer multiples of the fundamental frequency. For a frequency $f$ the harmonic series is *f*, 2*f*, 3*f*, 4*f*, ... (for example 110 Hz, 220 Hz, 330 Hz, 440 Hz), with each successive harmonic of the laryngeal signal being weaker than its predecessor, until no more are detectable.

The vocal tract can be modelled as a tube with one end closed by a vibrating membrane, like the reed in a clarinet, the *source* of the harmonic signal, providing the basic melody of speech. The harmonics of this melody are modified by a *filter* to form different sound qualities, by changing the diameter of the tube at different places, mainly by means of tongue and lip movements. These changes in the filter change the strength of the different harmonics relative to each other. Frequency regions where harmonics are stronger in relation to other areas are termed *formants*. The formants determine differences between voiced sounds such as vowels, glides, liquids, nasals and voiced consonants.

In general, the study of both speech production and speech perception require expensive specialised otolaryngeal instruments for inspecting the articulatory organs, and the middle ear and the inner ear are not accessible without specialised imaging procedures. Some of the tools for articulatory analysis are easily acquired by phonetics laboratories, but others, including tools for auditory analysis, are available only to qualified ear, nose and throat specialists. A great deal of information about these issues is available via internet search for relevant terms and does not need to be detailed here.

Fortunately for researchers who do not have access to expensive or restricted measuring facilities, a main area of relevance of articulation and perception for linguistics and psycholinguists involves speech transmission, the interface between sound production on the one hand and sound perception on the other. Part of this interface area is dealt with in this tutorial. There are very many more issues which can be studied in the domain of speech transmission, both experimentally with laboratory speech recordings and using authentic speech corpus materials: discrimination between different properties of speech signals, or the conversion of continuously changing speech signals as into discrete linguistic categories during perception (a process known as *categorial perception*). Tatham and Morton (2011) and references there provide a wealth of further systematic information on speech production and perception.

### 3.2 A note on acoustic phonetics and speech corpora

The best understood phase in speech communication is the transmission phase, through which acoustic signals from any source pass, whether speech or not. The speech transmission phase extends from the lips through some arbitrarily complex physical or electronic medium to the ear, and is studied in *acoustic phonetics*. Models of speech transmission are developed either by traditional auditory observation and transcription, or by recording the speech signal, followed by transcription and annotation, and supported by computer visualisations of various transformations of speech signal (mainly waveform, spectrogram and F0 trace). Acoustic models are also needed in articulatory and auditory phonetics, but it is generally the study of the speech signal in the transmission phase which is known as acoustic phonetics. The acoustic phonetic domain is the most accessible domain for the 'ordinary working phonetician', thanks to the availability of free software for phonetic analysis, and for this reason it is the focus of attention in this tutorial article.

For acoustic phonetic studies, s*peech corpora* (systematic collections of speech data) are required. The following list contains basic informal guidelines for making speech corpora.

1. A *speech corpus* is a set of speech recordings together with *transcriptions*, *annotations* and *metadata*, often with basic *statistical analyses* of the data, conforming to certain *standards* which make it searchable, sharable, and interoperable between different hardware and software environments. A set of recordings or transcriptions alone is not a speech corpus in this sense.
2. There are many kinds of scenario for the recording of speech data, from the constructed contexts of *laboratory speech*, with specially prepared lists of words or sentences or *faux* dialogue scenarios which may or may not be realistic, to *authentic speech* in the form of recordings of spoken interactions in the wild, and broadcast or internet media. The specification of data type and of instructions required for gathering the data require careful preparation.
3. If possible, recordings should be made with high quality recording equipment, with a minimal sampling rate of 16 kHz, with the microphone about a span (the distance between outstretched thumb and little finger, about 20 cm) from the mouth, and positioned to the side to avoid wind and breath noise. Speakers ideally need to sip water every 5 or 10 minutes in order to prevent the vocal tract from drying as air passes through it during speaking.
4. Recordings need to be archived systematically, with systematic and informative filenames, preferably together with the data, and with transcriptions and annotations, and always with the metadata of who was involved, and in what role, what was recorded, how it was recorded, and where and when it was recorded.
5. Annotations are transcriptions with time-stamps linking them to the recorded speech signal, made with specialised software such as *Praat*, *Transcriber*, *WaveSurfer*, *Annotation Pro*. Typical speech categories for annotation on different tiers (parallel streams) are the phoneme and syllable (in a phonetic transcription alphabet), word, phrase and sentence (typically in orthography). A corpus with several kinds of prosodic annotation is the Aix-Marsec corpus (Auran et al. 2004). Phonetic and phonemic transcriptions are based on the International Phonetic Alphabet, but in practice, especially in speech technology, are often made in the keyboard-friendly SAMPA alphabet and its extension the X-SAMPA alphabet, in which the IPA is represented equivalently by standard ASCII (Unicode Basic Latin) keyboard characters.

6. The metadata of a speech corpus can be compared with library information data for books: information about the recording situation and equipment, the kind of speech (e.g. laboratory speech, word list, sentence sets, story, interview, authentic unscripted conversation). Documentation of the speaker's consent is needed. The ethical aspects of including personal metadata need to be considered and may require approval from an ethics committee.
7. The creation of speech corpora is time-consuming and therefore expensive, which motivates the standardisation of techniques for speech corpus creation, which in turn ensures sustainability, reusability and interoperability of the corpus. Detailed information is available in Mehler et al. (2012) and further references there.

Speech corpora already exist for many languages, for example the CASS corpus for Mandarin Chinese (Li et al. 2000) and the Aix-Marsec corpus for English (Auran et al. 2004), which (among others) are used in this tutorial, and many such corpora are available in local and international data repositories. The number of corpora is rapidly increasing, enabling the use of powerful statistical discovery algorithms ('machine learning') to support the analysis of speech sounds.

The automatic extraction of information from the annotations in speech data is known as *annotation mining* (Gibbbon and Fernandes 2005) or *markup mining*. Two basic types of information can be extracted automatically from the transcription text in the annotation and its associated time-stamps, using relatively simple scripts written in programming languages such as the Praat scripting language (Boersma 2001) or Python. Starting from the annotation, detailed statistical analyses of speech timing and, with the annotation as an index for the signal, analyses of the speech signal itself can be generated automatically (Gibbon and Yu 2015; cf. also the example analyses in this tutorial).

### 3.3 Speech transmission: fundamental frequency

Frequencies are generally measured on the linear Hertz (Hz) scale, on which 1 Hz is defined as one cycle of the signal amplitude per second. The relevant frequency range for human audio perception, depending on sex and age, is from about 60 Hz to about 20 kHz. Frequencies above about 22 kHz count as ultrasound. The frequencies which are relevant for the phonetic measurement of tone and intonation patterns range from about 70 Hz to about 250 Hz for male voices, and from about 120 Hz to 350 Hz for female voices, and up to 500 or 600 Hz or more for small children and babies. These ranges are common everyday ranges with unexcited speech and are not absolute limits. For example, the female newsreader in Figure 16 below starts her paratone episodes with unusually high frequencies above 400 Hz. With very strong emphasis or excitement, or in screaming, even male and female adult voices may reach frequencies far above 400 Hz.

There are differences of opinion in different disciplines about whether *fundamental frequency*, the lowest frequency in the harmonic series of a periodic signal, should be abbreviated as *F0*, or *f0* or $F_0$ or $f_0$, meaning a zero formant or a lowest frequency, or as $f_1$, meaning the first harmonic. F0 is chosen here, following best practice in manuals such as the *Praat* manual. Some phoneticians pronounce F0 as /ɛf ˈoʊ/, but this is idiosyncratic, and potentially ambiguous, and originates in conventions for pronouncing 'zero' as 'oh', for example in English telephone numbers.

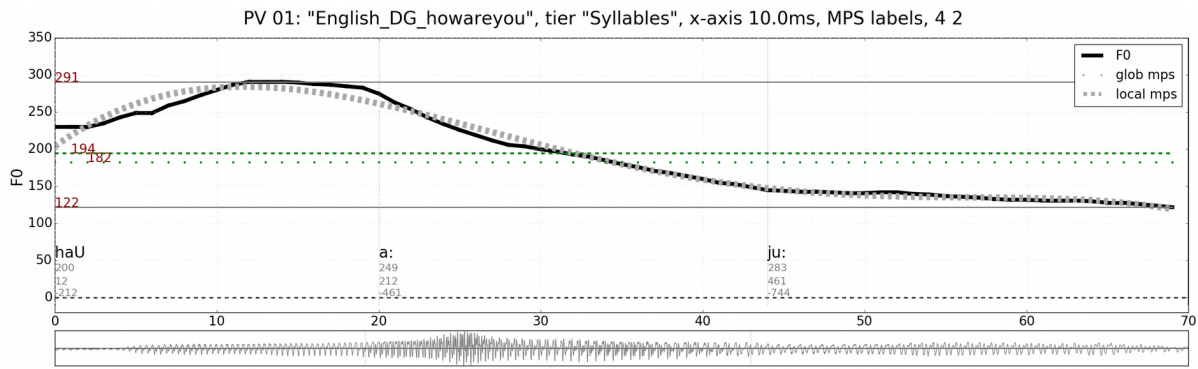

Figure 4: F0 trace for the utterance "How are you?" (male voice), with phonemic annotation in SAMPA symbols, with superimposed model.

Figure 4, displays the fundamental frequency for a recording of the utterance "How are you?", with a fairly emphatic *are*, and introduces the kind of visualisation which characterises most of the figures in this tutorial. The utterance shown in Figure 4 is quite short and the pitch accent is hardly distinguishable from the overall contour, which is falling (or high-low, H-L). The stress position for *are* is marked mainly by the duration of the [a:] vowel interval (249 ms) in relation to the duration of the preceding *how* (200 ms). The final *you* is still longer (283 ms), a pre-pausal lengthening effect. In functional terms, Figure 4 shows a typical F0 contour for a *who-which-when-where-how* type of information question (*wh-question*) in English, which tends to occur with a falling F0 contour. In more addressee-oriented, turn-eliciting *wh*-question utterances, rising final F0 contours also occur. Many other types of question occur with a rising contour, though these, too, may be overridden by discourse rank personal and social constraints and be pronounced with falling pitch.

The recording and TextGrid annotation input visualised in Figure 4 is the same as in Figure 1, but Figure 4 is generated the *PV* visualisation tool. From a direct observational perspective, the utterance starts fairly high for a male voice at about 230 Hz, continues with a rising pitch, reaches the peak of 291 Hz at the end of *how*, and gradually falls to the final frequency of 122 Hz at the end of *you*. The vertical *y*-scale of Figure 4 is the F0 scale. Horizontal lines through the F0 trace show the highest frequency in the recording (291 Hz, top line), the lowest frequency (122 Hz, bottom line), the mean frequency (182 Hz, dashed line) and median frequency (194 Hz, dotted line). Numbers on the horizontal *x*-scale indicate the sample frame number: each frame is 10ms long, so the duration of the signal in the figure is just under 700 ms or 0.7 seconds long. The signal sampling rate (44.1 kHz) is not indicated.

In addition to the F0 trace, a smooth stylised model of the F0 curve based on a polynomial approximation to the F0 contour is shown in thick dashed lines, and is explained below. The text labels shown in the Figure are derived from the TextGrid input to *PV*, and so are the durations of the label intervals and the time stamps. Below the F0 graph, a small graph of the original waveform is shown for orientation purposes.

The physical speech signal represented in Figure 4 was converted into a digital signal by sampling at regular time intervals. The digital sampling frequency (sampling rate) must be at least twice the frequency of the highest frequency to be measured in the signal in order to be sure of faithfully capturing a full cycle of the highest frequency and avoiding artefactual lower frequencies, known as *aliasing*. The condition $f_{sampling} > 2f_{signal}$ is known as the *Nyquist theorem* (also as the Nyquist-Shannon sampling theorem or the sampling theorem), and $2f_{signal}$ is known as the *Nyquist frequency*. For this reason, sampling frequencies for high quality audio are often 44.1 kHz (the CD standard) or 48 kHz (a digital tape standard), because the upper limit of hearing in healthy youngsters is around 20 kHz. For measuring the frequencies of speech, sampling rates of 22.05 kHz and 16 kHz are common; these lower frequencies are usually fine for phonetic analysis because the frequencies which are important for speech perception are below about 8 kHz. A frequency of exactly 44.1 kHz may seem a little bizarre, but it dates back to times when signal processing hardware and software was much slower, and conveniently provided a wide range of useful factors for easily converting to lower sampling frequencies: it is the square of the product of the first four prime numbers, or, spelled out in detail: $2\times2\times3\times3\times5\times5\times7\times7 = (2\times3\times5\times7)^2 = 44100$.

The Hertz scale is not the only frequency scale used in phonetics. Human hearing is not linear like the Hertz scale, but approximately logarithmic: frequencies are perceived as being equally spaced if they are related by ratios, not by addition. So the frequencies in the series *f*, 2*f*, 4*f*, 8*f*, ... i.e. $f \times 2^n$, are, simplifying a little, perceived as being equally spaced, unlike the frequencies in the harmonic series *f*, 2*f*, 3*f*, 4*f*, ... i.e. *f*×*n*. In order to model the hearing-related frequency series, logarithmic and logarithm-like scales are often used, relative to some base frequency, e.g. 50Hz.

The most commonly used scale of this kind is based on the musical interval of the *tempered semitone*, relative to some base frequency, for example, 50 Hz. There are 12 tempered semitones in an *octave*., which is a ratio of 2:1. The octave of a frequency *f* is 2*f*. The *tempered semitone* interval is a ratio which is one twelfth of an octave, i.e. 1.059:1 ($1.059^{12} = 2.0$). The *just semitone* derived from natural musical scales, is not used, because different positions in different scales lead to different definitions of the just semitone, for example 25:24 (1.0417:1) or 16:15 (1.066':1). Frequency ratios are referred again later to in connection with chanted contours in the context of discourse prosody. In this tutorial introduction, the linear Herz scale is used. There are other logarithmic and near-logarithmic scales which are often used in phonetics because they are better models of the way human hearing is structured.

There are several methods for estimating the fundamental frequency from a speech signal. Before further processing, the signal is filtered to weaken harmonics or noise frequencies which are lower or higher than the desired frequency range. Voiceless sounds such as voiceless consonants or whispers are not periodic, and therefore do not have a fundamental frequency in this sense, though an impression of frequency may be perceived. The basic idea is to travel along the samples of the filtered digital speech signal in steps, looking at the signal through a two-dimensional *moving window*, often of length 10 ms or 20 ms, which not only selects samples in intervals defined by the window length, but may also filters these samples further in order to reduce certain types of distortion. The technical term for fundamental frequency extraction is *fundamental frequency estimation*, since different window lengths and filters, and different analysis procedures may lead to somewhat different results. The analysis procedures use either *time domain analysis* (for example measuring the length of a cycle) or *frequency domain* analysis (for example measuring the distance between harmonics). For further details on F0 estimation Hess (1983) or Gibbon and Richter (1984) can be consulted. The pitch estimator algorithm used in the *PV* software, *RAPT* (*Robust Algorithm for Pitch Tracking*) was developed by Talkin (1995).

## 3.4 Perturbations, harmonics, formants

Consonants affect the vibration of the vocal folds by narrowing the vocal tract and by obstructing the flow of air in the mouth, changing the air pressure at the vocal folds, so that the frequency at the start of the consonant becomes lower when the airflow slows down, and increases when when the consonant is released and the airflow accelerates. The extreme case is the voiceles stop, resulting in downward spike at the beginning of the consonant, the disappearance of the F0 trace while the vocal tract is closed, and a restart from a relatively high frequency down towards the reference line. The effect with other obstruents generally not so obvious. The very short patterns resulting from the modification of F0 by consonants are known as *consonantal perturbations* or *pitch perturbations*, and, together with minor F0 modifications associated with different vowels, are known as *microprosody*.

An F0 trace with *PV* or Praat shows the perturbations clearly. Figure 5 shows a *PV* representation of a recording of two interpausal units: the two English sentences *Peter Piper picked a peck of pickled pepper, Mary Miller moaned on Monday morning*. In addition to the F0 trace, two polynomial curves are shown: the first globally across both sentences, and the second locally across the voiced sections of the interpausal units. In the recording of the first sentence, lexical items all start with voiceless consonants, while in the second, there are no voiceless consonants at all. The differences in the F0 traces of the two sentences are very conspicuous. In the second sentence the F0 proceeds rather smoothly and is modelled locally as a continuous and quite well-fitted smooth curve across the second interpausal unit.

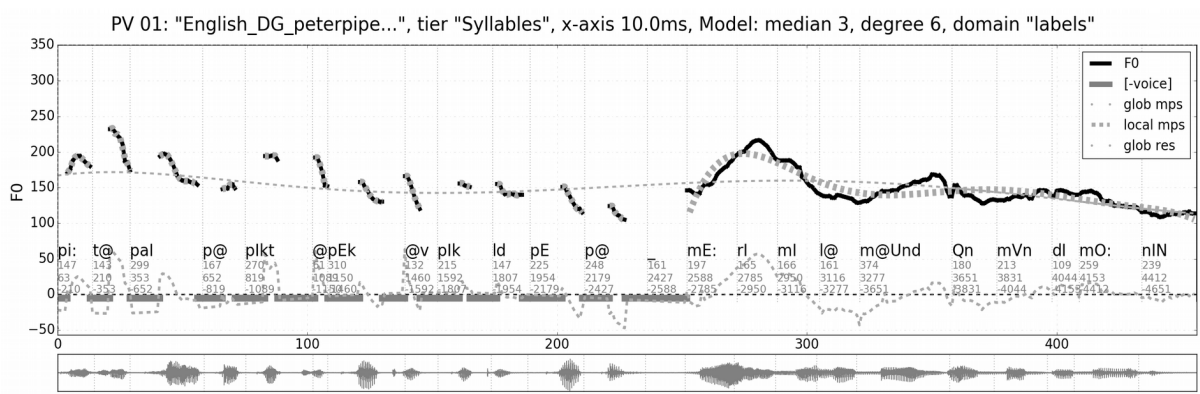

Figure 5: F0 model and residuals for "Peter Piper picked a peck of pickled pepper. Mary Miller moaned on Monday morning." Model: thin dotted; residual: thin dashed.

This smooth modelling is not possible with the first sentence: not only do the voiceless consonants create breaks with no F0 between the vowels, but, depending also on other factors such as the stress position pattern, the F0 is lowered before each voiceless consonant and jumps to a higher level after consonant before falling to the expected level on the vowel and falling again before the next voiceless consonant. In Figure 5 these consonantal perturbations are modelled as the differences between the smooth polynomial model of the F0 contour (*residuals* of the polynomial model) and the rough F0 contour itself, and are shown as curve which is similar to the F0 curve, but centred on the zero line. There are two main kinds of property of this residual contour which differ from the original F0 measurements:
1. Consonantal perturbations: short, sharp, sudden movements in F0.
2. Pitch accent contours: much slower and either fairly smoothly rounded, as in the second sentence, or dwarfed by the sharp, sudden movements of consonantal perturbations, as in the first sentence.

The obstruent consonants (plosives or stops, fricatives, and affricates) generate additional noise frequencies because of friction caused by air being forced through the narrow gap between upper and lower articulators in the vocal tract: teeth and lower lip or tongue, roof of the mouth and tongue, pharynx and tongue, larynx. The different noise frequencies associated with different consonants can be illustrated by producing and listening to the voiceless fricatives [f], [ʃ], [ç], [s], ranging from the lower frequency noise of [f] to the higher frequency noise of [s].

The harmonics generated together with the fundamental frequency are filtered in the vocal tract tube, as already noted. The frequency ranges in which the harmonics are stronger are called *formants*. The range of harmonics which is relevant for the fine-grained perception of vowels, covering the first three formants, lies between about 300 Hz and 3000 Hz. For example, when the tongue and lips are in a neutral central position in the vocal tract tube and the neutral vowel [ə], schwa, is produced, in which the vocal tract tube has a neutral position and , in the ideal case, the first three formants are approximately equally spaced. The exact frequency of the formants depends on the length of the vocal tract from the vocal folds to the lips.

Figure 6 visualises a schwa [ə] spoken with a fundamental frequency (F0, the lowest dark horizontal area) of about 140 Hz and three typically evenly spaced schwa vowel formants at about 600, 1600 and 3600 Hz. In nasal sounds, the lowered velum opens the nasal cavity, adding an additional lower frequency nasal formant. The repeated opening and closing cycles of vocal fold vibration are shown in the upper section of Figure 6. In the lower section the vertical pattern shows the phases of vocal fold opening (vertical dark stripes) and closing (light with horizontal stripes), and in the middle the three equally spaced horizontal stripes show the formants of the vowel [ə]. The top horizontal dark area shows higher formants.

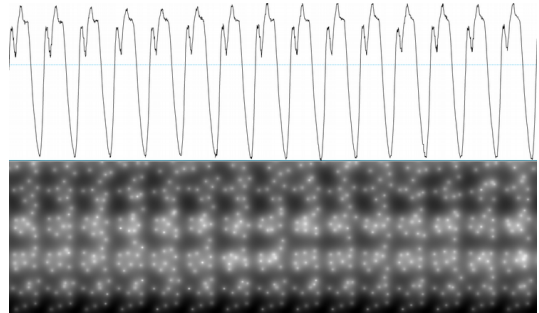

Figure 6: Praat screenshot of waveform (upper) and spectrum (lower) of [ə], showing 1ms intervals between peaks, corresponding to a fundamental frequency (F0) of about 140 Hz.

The same recording as the one shown in Figure 4 is visualised in the two Praat screenshots in Figure 7. The screenshots show two spectrographic displays of the frequencies composing the signal. The narrow band spectrogram display on the left shows the equally spaced harmonic series derived from the F0; the F0 itself is the 'zero' formant or first harmonic, at the bottom of the display. The F0 contour has the same shape as the harmonics, and is superimposed on the spectrogram using its own frequency scale. In the narrow band display, the frequency resolution is more fine-graned, showing the first ten or so harmonics of F0 very clearly as a series of horizontal wavy lines, while the temporal resolution is blurred. The formants are shown as areas where harmonics in some frequency ranges are darker.

In the broad band display, on the other hand, the opposite relation between time and frequency is found: the frequency resolution is blurred, which brings out the formants and blurs out the harmonics, while the temporal resolution is so fine-grained that the effects of the individual openings of the vocal folds are shown as vertical lines. The formants are shown as darker areas, and are entirely parallel to the formants in the narrow band display.

The differences in frequency of the formants determine the quality of the vowels and sonorants: in the middle of the graph, for example, the first two formants of [aː] are so close together that they practically merge, while on the [j] they are far apart, conditioned by the different shapes of the vocal tract tube.

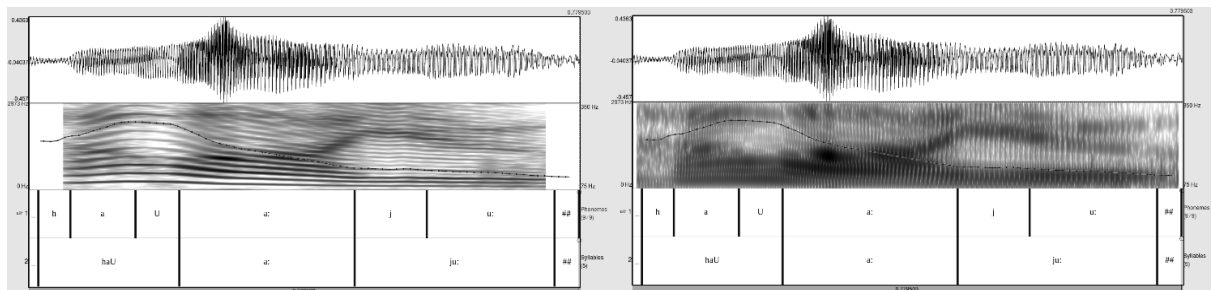

Figure 7: Praat F0 trace with narrow band (left) and broad band (right) displays for recording of "How are you?".

## 3.5    Visualisation and stylisation

The visualisations in Figure 5, Figure 6 and Figure 7 demonstrate different techniques for transforming selected properties of the speech signal into the vision modality in order to make the structure of speech signals intuitively easier to understand, and to support insight into patterns of speech which can then be further investigated by other visualisations or by quantitative analysis of corpora, or by experimental methods. Systmatic visualisation in scientific research is a rapidly growing discipline in applied mathematics and computer science, and various basic types of visualisation are used by any phonetic software tool, including the *PV* software in the present tutorial. The three basic types of visualisation are the waveform, the F0 trace and the spectrogram. The *PV* software does not show the spectrogram, but focusses on properties of F0, including *stylised models* of the F0 trace. For example, the visualisation in Figure 5 shows the F0 trace as a continuous dark line, and a stylisation which represents F0 change over time in a simplified but principled way. A very simple and frequently used kind of stylisation is *median filtering*, in which the statistical median of sequences of 3, 5, 7 etc. F0 measurements in successive segments of the speech signal (*windows* over the speech signal) are calculated. The median filtering procedure removes extreme values of the speech signal, such as those caused by consonantal perturbations, or errors in the F0 algorithm.

Many other approaches to the computational modelling or stylisation of F0 contours have been proposed both for phonetic and linguistic theory and for speech technology applications, particularly in text-to-speech synthesis. Among many milestones in prosody modelling for speech synthesis perhaps the most influential are: the Fujisaki model of phrase and accent components of intonation (Hirose et al. 1984); the linear component model of 't Hart et al. 1990; the Tilt model of Taylor (2000). Computational models of intonation and tone, both based on finite state machines, were developed by Pierrehumbert (1980) and Gibbon (1987, 2001). Hirst's *Momel* model (Hirst et al. 2000), Mertens' Prosogram model (2004) and the collection by Sagisaka et al. (1997) represent experimentally based computational phonetic models of prosody, particularly F0 patterning. The development of research on F0 modelling and stylisation can be followed in the proceedings of the various conference series, particularly *Speech Prosody* (*SP*) and *International Congress of Phonetic Sciences* (*ICPhS*).

In the present tutorial context, an approach to stylisation is taken which is an extension of approaches which are sometimes used in experimental research on fundamental frequency patterns. A smoothed line representing F0 is generated over a domain or a combination of domains such as the syllable, the word, the phrase or the utterance, by means of simple or multiple linear regression analysis. The simplest form is the straight line which best fits the sequence of F0 measurements, generated by *simple linear regression* analysis. The simple linear regression function can be expressed as $y=a+mx$, where *a* represents a baseline where zero value of *x* meets the *y* axis, and *m* represents the slope of the line (*m* can be thought of as a mnemonic for multiplication, and *a* for addition).

In general, simple linear regression is too simple to represent the F0 pattern in the syllable, word, phrase and utterance domains, though it is sometimes used, and the fundamental idea of a baseline which is added to a sloping pattern is a helpful basic model for F0 patterns. The F0 pattern is represented better if different values of *a* and *m* are combined in *polynomial regression* (*multiple linear regression*) of higher *degree*:

$$y = a_0 + a_1x^1 + a_2x^2 + a_3x^3 + \ldots + a_nx^n$$

For some of the figures in this tutorial a polynomial approximation of degree 2 is used, i.e. $y = a_0 + a_1x^1 + a_2x^2$, but in some cases, such as Figure 5, a higher degree is used in order to capture more variation in the F0 curve. An empirical 'rule of thumb' used for determining the degree of the required polynomial was that the degree is equal to or greater than the number of items (e.g. pitch accents) which modify the global trend. In Figure 5 both the global and local models have degree 6 (documented in the figure title).

If the degree of the polynomial is too high, the resulting curve is too close to the original F0 measurements to be of much use for gaining insights about the structure of the F0 curve, though it may represent an ideal to be achieved in speech synthesis. The criterion for fitting the polynomial curve to the data is 'the best fit', which is the lowest sum of squared differences between the model and the original, point for point (*least squares* method). The approach taken here is called the *MPS* (*Multiple Polynomial Stylisation*) model, because stylised models can be created simultaneously over different time domains (for instance the whole recording, or words, or syllables, as shown in Figure 5 and in several other figures.

There are many other kinds of stylisation, developed with different objectives in mind. The Fujisaki model mentioned above, for example, is explicitly based on a model of speech production, while Mertens' Prosogram model (2004) is explicitly based on a model of speech perception. The MPS model is neutral with respect to speech production and perception, though it is open to interpretation in both of these domains.

# 4 Specific cases: lexical prosody

## 4.1 Mandarin Chinese tones in citation contexts

Perhaps the most well-known example of lexical tone is Mandarin Chinese, with four lexical tones (shown in Figure 8), and a neutral, context-dependent fifth tone. In Figure 8 the four lexical tones are shown with the syllable [ma] uttered in isolated citation contexts. A more sophisticated interpretation of the meanings of *ma* than one usually finds was kindly provided by Li (2017) (the 'kind of fibre' in the quotation is usually glossed as 'hemp'): "For tone 1, firstly it could mean 'mother' (妈); secondly it means 'clean sth' (抹). These two are different characters in Chinese. For tone 2, firstly it means 'numb';

secondly it stands for a kind of fibre. Two meanings but with the same Chinese character (麻). For tone 3, firstly it means 'horse' (马); secondly, it could mean a measurement unit which is equal to yard (码). For tone 4, it means 'scold sb' (骂)."

Two polynomial models of degree 4 are superimposed: over the whole sequence and also over the individual syllables (which are monosyllabic words). In the case of tone 2 and tone 3, the annotation boundaries are a little indeterminate, and the F0 trails off a little after the end of the annotation of the perceived syllable. The global model shows the *ordering effect* which arises when different citation forms are grouped in the same utterance, where the prosody of the citation forms is influenced by the overall intonation of the utterance, as if examples were elements of a list (higher or rising F0), or the closing boundary tone (lower or falling F0). In the present tutorial context this is deliberate, but must be avoided in experimental scenarios by an appropriate instruction to the speaker, and by using several examples of each citation form and random ordering of the examples.

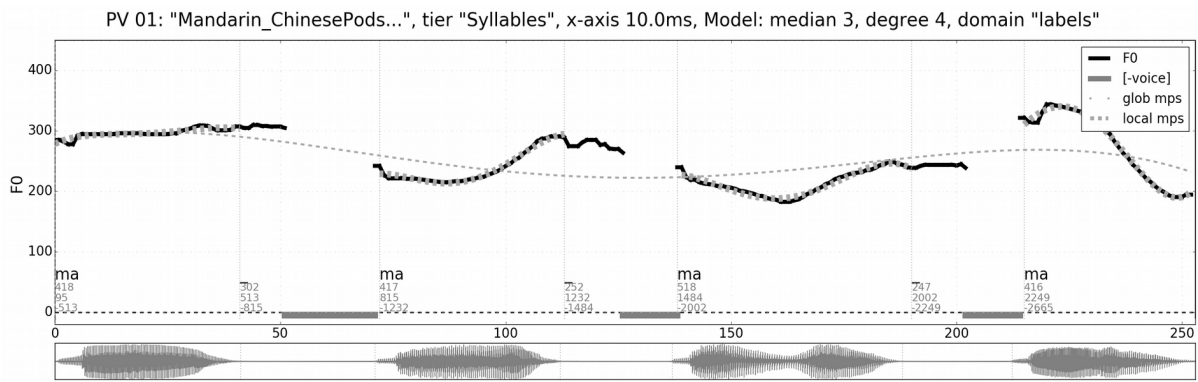

Figure 8: Mandarin tones on syllable "ma": 1 high; 2 rising; 3 falling-rising; 4 falling (female voice).

In addition to the tones, other noteworthy properties of the Mandarin lexical tone system are visible. The first property is the use of *creaky voice* to emphasise or replace very low F0. Figure 9 shows creaky voice in two different contexts in which low frequencies occur. The first creaky voice context is seen in the the utterance final positions at the ends of tone 2 and tone 4. This phenomenon arises because a citation context is not just a word but also an utterance at the same time.

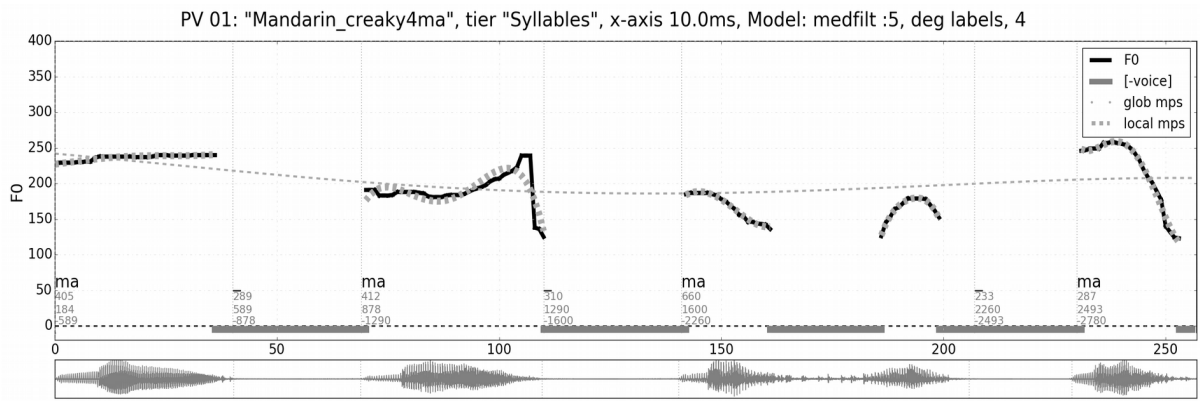

Figure 9: Mandarin tones on the syllable "ma": 1, high; 2, rising; 3, falling-rising; 4, falling (female voice), with creaky voice at the ends of tone 2 and 4, and in the centre (at the low F0 point) of tone 3.

In Figure 8 the waveform shows a distinct decrease in amplitude in the middle of tone 2. This is exactly the second context for creaky voice shown in Figure 9. The creaky voice in this context appears as a gap in the F0 trace, and as fairly regular pulse groups in the waveform graph. In the pre-pausal and tone 3 contexts contexts is a very strong and regular, frequently observed tendency in Mandarin Chinese, and indeed may be observed to occur over whole utterances in some discourse contexts.

Mandarin Chinese tones also vary phonetically in the way in which the tones affect their neighbours in sequences of syllables, with assimilations and replacements which depend on the neighbouring tone. The process which results in these "allotones" is sometimes called *tone sandhi*, or *tonal sandhi*, after a Sanskrit word for contextual variation of sounds. Tone sandhi is not manifested in monosyllabic citation forms of the kind shown in the Figures. A different kind of tone sandhi, which is characteristic of Niger-Congo tone languages, reappears below in discussion of the Tem language. Extensive further information on tone is provided by Yip (2002).

## 4.2 Tones of Anyi/Agni, Niger-Congo, Kwa (ISO 639-3 *any*) in citation contexts

A majority of African languages belong to the Niger-Congo language family and have tonal systems which are quite similar to each other in general terms, but very different in detail. The *register tone* or *level tone* systems differ considerably from the *contour tone* patterns found in Mandarin Chinese. Many Niger-Congo languages have just two 'register tones' or 'level target' tones, high (H) and low (L). The tones are associated with syllables and subject to various production constraints, for example tone spreading over unassociated yllables in longer domains such as the word, to dumping or 'crowding' of more than one tone on one syllable, and to other processes, including conditioning of the tone in various ways by consonants.

The Kwa language Anyi (or Agni in the French spelling of the name), Ivory Coast, in the Niger-Congo group, provides a clear example of some of the main features of Niger-Congo register tone systems (Adouakou 2005). Figure 10 shows two words distinguished by tone: *ánónmàn* (H-H-L) meaning yesterday, and *ànònmǎn* (L-L-LH) meaning bird or grandchild.

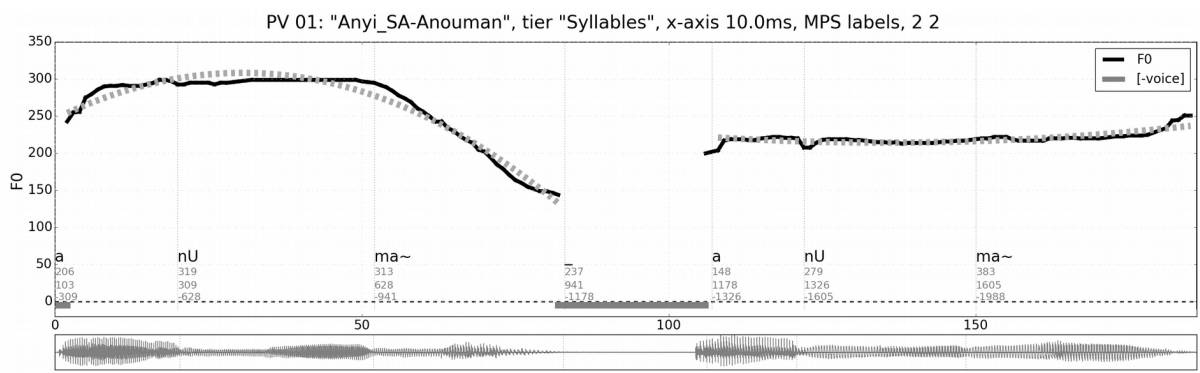

Figure 10: Falling and rising tonal contrast on Anyi (Niger-Congo, Kwa) "anouman", with polynomial model superimposed.

## 4.3 Tones of Kuki-Thadou, Sino-Tibetan (ISO 639-3 *tcz*) in citation contexts

The tones of Kuki-Thadou (Haokip 2015), a Sino-Tibetan language of North-East India, have received different phonological analyses, partly owing to analysis of slightly different dialects and partly due to different theoretical positions (Gibbon et al. 2009; Hyman 2010). The tones are described here as H (HL in Hyman's account), LH and L (high, low-high and low). In these analyses the citation context tones were not controlled for the ordering effect of utterance finality, which may also account for differences in analysis. In addition to the tone inventory analysis, Hyman also identifies contour simplification and tone spreading processes in Kuki-Thadou which are not dealt with here, noting that the properties of Kuki-Thadou are of a kind which is more typical of African tone languages than of Sino-Tibetan tone languages.

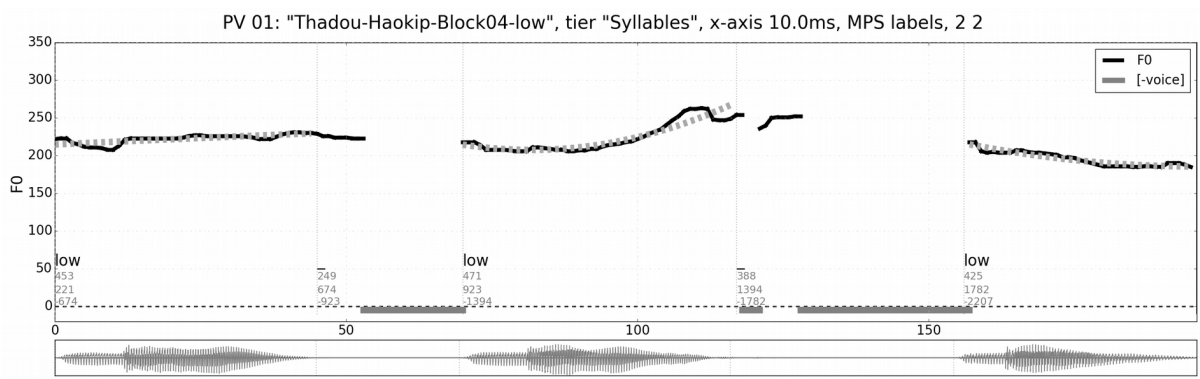

Figure 11: Kuki-Thadou (Sino-Tibetan, North-East India) tones on the syllable "low": high tone, rising tone and low tone (female voice).

In Figure 11 the F0 patterns for *lów* (H, field), *lǒw* (LH, medicine) and *lòw* (L, negative marker) are shown in a citation context. The acute, haček and grave diacritics on [ó ǒ ò] stand for high, rise and fall, respectively, following Africanist rather than Sinologist conventions. The H tone in Figure 11 has mean 222 Hz and median 223 Hz respectively and a very small rising linear regression slope of 0.34. The L tone has mean 196 Hz and median 194 Hz, with a very small falling linear regression slope (-0.77). The LH tone has a lowest F0 of 206 Hz and highest of 263 Hz, with an expected larger linear regression slope (1.14). The differences between the tones are clear, though the names of the tones might be disputed; the low term could be categorised as a falling tone, for example. Clearly, for a full account many more examples are required, and the operations which Hyman analyses need to be analysed further in context.

### 4.4 English stress positions in citation contexts

The prosody of words includes not only phonemic prosody, discussed in the previous section, but also tonal and accentual aspects of word structure. Both tones and stress positions may play a role in the morphology of word formation and the morphology of inflection, for example. Tonal morphology is found in Niger-Congo languages, for example in Ibibio where a high tone links compound names, and a contrast between high and low tones may indicate distal temporality, with high tone indicating future temporal distance or past temporal distance, and low tone indicating future or past temporal proximity. The present subsection deals only with aspects of lexical stress positions and their association with pitch accent in English. Semantic and syntactic aspects of lexical stress positions are not discussed.

English has the second type of pitch accent, as already noted, with variable F0 realisations in which the F0 pattern on the accents may be high, low, rising, falling, rising-falling-rising, falling-rising-falling, and may be merged with initial and final F0 boundary markers. The choice between specific pitch accent shapes is motivated by discourse semantics, such as information structure, and pragmatics, such as speaker or hearer orientation, rather than by grammatical structure. The relevant structural feature at lexical and phrasal ranks is therefore not the pitch accent melody itself, which is a phonetic means to a semantic or pragmatic end, but the stress position associated with the pitch accent pattern.

It was noted in the introductory section that the terms used for the positional feature of the pitch accent system are *stress*, *stressed* (syllable, word, phrase), and *stress position*. Accordingly, English, and related languages such as Dutch and German, are sometimes referred to as stress-intonation languages or stress-accent languages. The two main markers of stress position in English are pitch accents and longer duration of stressed syllables. This is a statistical tendency rather than a rule because syllable duration is determined by several other factors involved: phonotactic complexity, long-short vowel contrasts, shorter vowels before voiceless consonants, longer vowels in voiced environments, and intrinsic vowel duration.

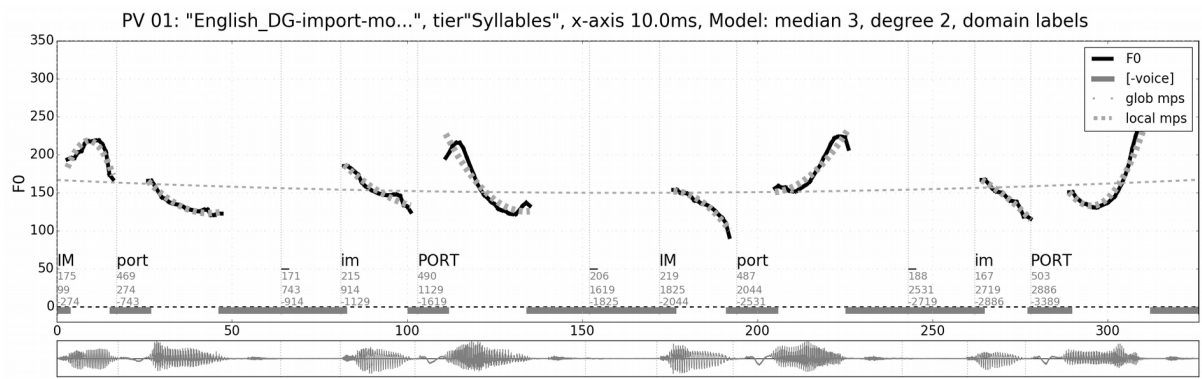

Figure 12: English 'import' with initial and final stress, in declarative (falling intonation) and interrogative (rising intonation) contexts. Model: thick dashed.

Lexical stress position is a key phonological and morphological feature in English lexical items like *TElephone*, *teLEphony*, *telePHOnic*, but specific phoneme-like contrasts in simple lexical items of the *INsight-inCITE* type are quite rare. Structure marking occurs in morphological contexts, such as in nominal phrases combined with compound lexical items: *light HOUSEkeeper* as opposed to *LIGHThouse keeper*. Neutralisations may occur because of the override effect of higher ranked stress patterns, resulting in amusing ambiguities, a staple element of English humour: *ice-cream* and *I scream*, for example, as in the Dixieland jazz song: *I scream, you scream, everybody likes ice-cream*. Some semi-regular examples have already been noted, such as *IMport*, with stress on the first syllable, as a noun, meaning an imported product, as opposed to *imPORT* as a verb. In contrastive contexts, this distinction may also be overridden by a higher ranking discourse rank semantic or pragmatic constraint, as in the humorous example: *This whisky was not EXported, it was DEported*. Nevertheless, the stress pattern is an important property of English words, and incorrect stress, even if not phonemically distinctive, may lead to low intelligibility of lexical items. As with other prosodic patterns, these lexical patterns may be overridden by higher ranked discourse conditions.

The visualisation in Figure 12 shows the word form *import* in four different contexts, with each example characterised by an F0 break on the voiceless [p] (from left to right):
1. *IMport* (noun) in an *informative* context with falling intonation: the high pitch accent on the first syllable is well above the reference line provided by the model, falling in the informative context to a final low boundary tone.
2. *imPORT* (verb) in an *informative* context with falling intonation: the initial syllable is approximately equally above and below the polynomial reference line, while the high pitch accent on the second syllable is well above the reference line, falling in the informative context to a final low boundary tone.
3. *IMport* (noun) in a *questioning* context with rising intonation: the initial syllable of IMport is low in the context of an overall rising tone, rising to a high boundary tone in the questioning context.
4. *imPORT* (verb) in a *questioning* context with rising intonation: the initial syllable of imPORT is low in the context of an overall rising tone, with a low accent on the final syllable, rising to a final high boundary tone in the questioning context.

The time-stamps of the annotations in Figure 12 show that the stress position, and thus also the pitch accent at this stress position, is not always associated with a longer syllable. One of the factors conditioning duration in these examples is the phonemic length contrast in English: the syllable *im* has a shorter vowel than the syllable *port*, which co-determines the duration of the syllable. Another factor involved, since the contexts are citation contexts in which each word is simultaneously an utterance, is the pre-pausal lengthening.

# 5 Specific cases: phrasal prosody

## 5.1 Tem: tones in context

When lexical tones occur in context, their phonetic form is conditioned by neighbouring tones, in tone sandhi, by intonation and in some cases by the syllable patterns they accompany. In many Niger-Congo tone languages, the basic inventory has two lexical tones, high (H) and low (L), and sequences of these tones follow a regular rule: the H to L F0 movement is greater than the L to H F0 movement. This relation is a partial progressive assimilation of the second H to the L. In a sequence H-L-H, the second H is therefore lower than the first, conditioned by the intervening L, a relation called *automatic downstep*. The pattern produced by these conditions is called *tonal terracing* or *terraced tone*, because the visualisation of the overall contour has been said to resemble a sequence of relatively level terraces. There are terracing patterns which are more complex than this basic case. A computational model for these terracing relations has been proposed (Gibbon 1987; Gibbon 2001).

Sometimes the syllable with the intervening L between two syllables with H tones is deleted, replacing an H-L-H sequence by an H-H sequence: the second H is not at the same F0 level as the first H or perhaps a little lower, as one might expect, but much lower, conditioned by the underlying but phonetically deleted L, which remains as a *floating low tone*. The floating low tone may result from language change, for example disappearance of the syllable with which the tone is associated. If the tone also disappears, a *lexical downstep* may result with a sequence H-!H in which the second H is lower than the first. Further complications for tones in phrasal contexts are:
1. the tonal assimilation may be total, and apply to both H and L tones: a lexical H-L-L tone sequence may be produced in context as H-H-L, and a lexical L-H-H tone sequence may be produced in context as L-L-H;
2. the domain of lowering may be delayed, so that lexical H-L-L may be produced as phonetic H-L-H;
3. syllable structure, especially consonants, may block or change tone productions;
4. where there are more syllables than tones, the tones may spread to cover the unassociated syllables;
5. where there are more tones than syllables, they may be 'dumped' or 'crowded' on to one syllable, often the final syllable of the relevant domain;
6. a general principle may apply, ensuring that sequences of identical lexical tones do not occur but are spread from a single lexical tone during production where necessary, the *Obligatory Contour Principle*.

This variety of tone production effect is not discussed further in this context. The example from Tem in Figure 13 illustrates the main point, with a few examples of language-specific tone sandhi (data based on Tchagbale 1984).

Figure 13 shows the relation between syllables, lexical tone and F0 pattern on the sentence *kodóngariké nyazi wúro ta sí* (laughing as though the king had not died), where the acute accent represents H tone and L tone is not marked. Inspection of the F0 contour, taking unevenness due to microprosody into account, shows that there is a total progressive downstepping assimilation not only of lexical H to preceding lexical L but also of lexical L to preceding lexical H, except on the final syllable. The overall mapping is shown in Table 1, where ↓ is used for downstepped H tones and ↑ for upstepped L tones.

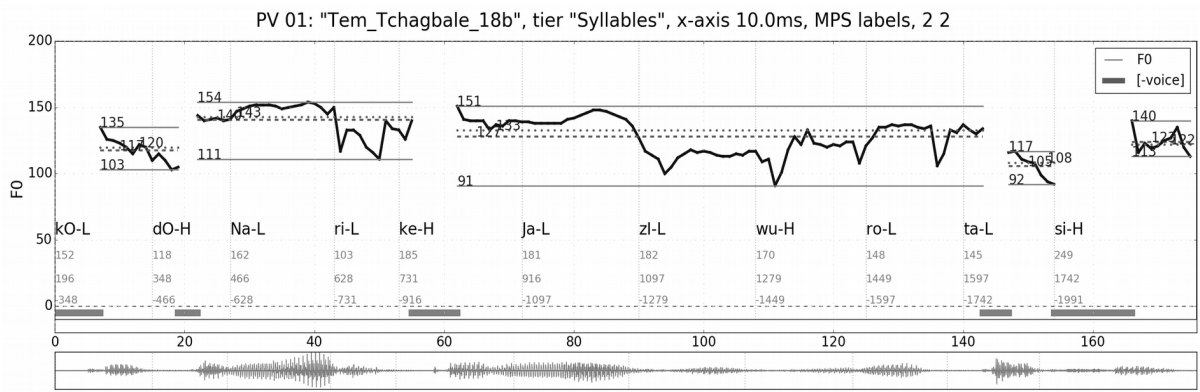

Figure 13: Tem (Niger-Congo, Gur) tone patterns in context: "kodóngariké nyazi wúro ta sí" (laughing as though the king were not dead).

*Table 1: Tone sandhi mapping in Tem.*

| Syllable: | ko | dó | nga | ri | ké | nya | zi | wú | ro | ta | sí |
|---|---|---|---|---|---|---|---|---|---|---|---|
| *Lexical*: | L | H | L | L | H | L | L | H | L | L | H |
| *Condition*: | L | ↓H | ↑L | L | ↓H | ↑L | L | ↓H | ↑L | L | H |
| *Phonetic*: | l | l | h | l | l | h | l | l | h | l | h |

In the phonological literature, downstep is often written with the exclamation point "!", dating from pre-Unicode days, but here downstep and upstep are noted with down and up arrows respectively for the sake of typographic clarity. Also for clarity, upper case letters are used for the lexically contrastive tones, and lower case letters are used for the phonetic tones, which are quite easily relatable to F0 values.

Note that the conditioning of neighbouring tones derives from the preceding *lexical* tone, not from the preceding produced *phonetic* tone. The relation between lexical tone and produced tone is fairly clear in this example, but individual examples are not enough for discovering the tone system, and this degree of clarity is not always evident. Many test utterances with examination of tone contrasts in citation contexts are needed in order to discover the relevant inventory of lexical tones.

For the computationally minded: formal models can be designed for recognising and generating these downstepping and upstepping tone relations (Gibbon 1987, 2001). These relations in Tem can be formulated as a simple finite state transducer with iteration in three loops over four transitions (which can also be represented by a transition diagram or matrix): <$State_H$,H,h,$State_H$>, <$State_H$,L,h,$State_L$>, <$State_L$,L,l,$State_L$>, <$State_L$,H,l,$State_H$>. The two single state loops define terraces, and the loop between the two states defines the overall terracing pattern. The transducer can be initialised and terminated at either of the two states. Equivalent phonological rules can also be formulated: H→h/H_, L→h/H_, L→l/L_, H→l/L_.

## 5.2 Phrasal contexts for English stress

One of the most well-known contexts for the variability of stress positions in English is the contextual rhythm constraint called the *Thirteen Men Rule*: when a word with a final stressed syllable is immediately followed by a word with an intial stressed syllable, the stress position in the first word may fall on an earlier syllable. The sequence consisting of *thirTEEN* and *MEN*, for example, is pronounced in a citation context not as *thirTEEN MEN* but as *THIRteen MEN*. In the terms of a constraint from poetic prosody a constraint could be formulated: *Avoid spondees!* The exact syllable target for the relocated stress position depends on other constraints, however. For example, the complexity of the syllable structure of the words is a factor: in a sequence *asSIGN WORK* the stress positioning is generally not *ASsign WORK*. Otherwise, the constraint applies regularly in citation contexts, but it is basically a tendency, and can be overridden, as expected, by discourse constraints. An overriding semantic contrast constraint could yield: *I said thirTEEN MEN, not thirTY WOmen*.

Some contextual constraints were briefly discussed in the introductory section. In such contexts, not only the lexical stress placement in the word, but also the overall pitch contour of a phrase or other larger unit of speech may be heavily influenced by discourse factors. In the introduction section, isolated citation examples of the sentence *Willy married Mary* in different contexts were noted. The following examples (with constructed *faux* dialogue oriented contexts in square brackets) show stress positions assigned semantically or pragmatically as *contrastive stress positions* or *emphatic stress positions*, respectively, overriding phrasal and lexical stress positions.

1. [*After a long engagement,*] *Willy married MAry*. In this case, an unsurprising piece of news is communicated, resulting in a neutral stress placement pattern and a neutral pitch pattern.
2. [J*ACK didn't marry Mary,*] *WILLY married Mary*. The topic of the sentence, *Willy*, is focussed because of the semantic contrast with *Jack* and the remainder of the sentence is subject to *deaccenting* or, more generally, *post-focus compression* (Xu 2011). The contrastive context may be explicit, or implied (necessarily the case in isolated citation contexts).
3. [*Willy didn't marry LIZZY,*] *Willy married MARY*. In this example, the contrast is on the main focus of the sentence, *Mary*. In cases of final stress position there may be little or no difference between a semantically contrastive or pragmatically emphatic pitch accent at that position and a neutral pitch accent.
4. [*I can't believe it,*] *WILLY married MARY*. This example is intended to express pragmatically determined stress positions associated with surprised emphasis rather than semantically determined contrastive stress positions.

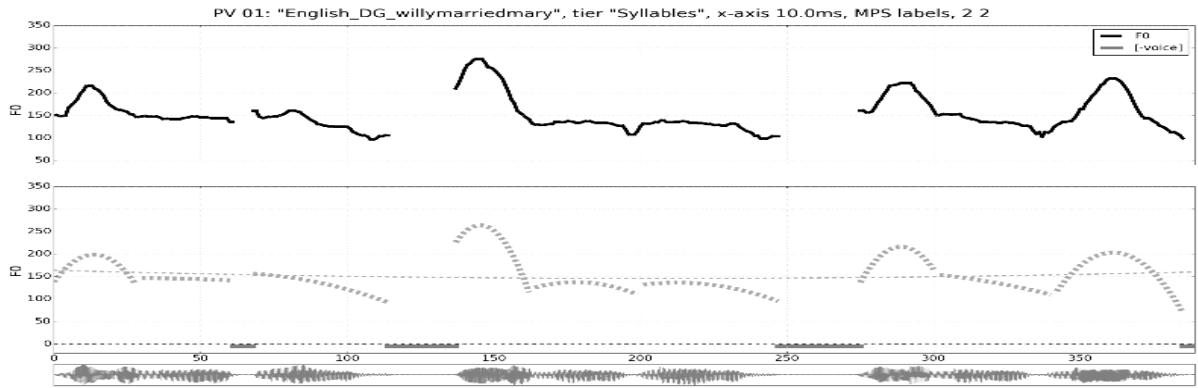

Figure 14: *Willy married Mary* in three different focal contexts (F0 trace, male voice).

Figure 14 (top) shows an F0 trace for *Willy married Mary* in three of the four contexts listed above: (1) neutral, (2) focussed semantic topic; (3) focussed semantic topic and focussed semantic comment. In Figure 14 (bottom) a stylised model of the pitch accent and intonation patterns is displayed without the original pitch contour, in order to clarify the differences between cases.

## 6 Specific cases: discourse prosody

### 6.1 Mandarin Chinese story reading

Discourse takes on very many forms: informal or formal, technical or casual, single-speaker monologue, as in story-telling or news-reading, or dialogues with short, everyday turns, or with lengthy turns, as in radio and television interviews. The present subsection illustrates two examples of reading aloud monologue discourse by Mandarin speakers, one male and one female. The object of these visualisations is not to show details of pitch patterning but to give an opportunity for holistic observation of the 'Gestalt' or coherent overall form, of a long text utterance with a hierarchy of parts. No annotation is shown. It is very rare to see long-term sequences of this kind (nearly 50 seconds for the male and nearly 60 seconds for the female) discussed in the literature, and reading aloud, an important form of communication, is a useful genre for this.

In Figure 15 the male voice is displayed at the top and the female at the bottom display. Both speakers read Mandarin Chinese translations of the traditional IPA documentary text, *The North Wind and The Sun*, which originates from the ancient Greek fables of Aesop. There is an element of compositionality in the F0 trace, with shorter local contours characterised by low F0 bandwidth (frequency range) and downtrending local F0 contours, which merge into longer, I don't think there will be a war in Korea. I know the situation there is very tense. But China government will work on it. If there is a war, China will be influenced. We do not like wars:).broader bandwidth and also downtrending global F0 contours to form a two-level flat hierarchy of iterative F0 patterns. The shorter local F0 units tend to have durations between 1 and 2.5 seconds for both male and female voices, and form the basic intonation units. The longer units into which the shorter local units merge, are paragraph intonation units or paratones, which mark episodes in the story. The paratones vary considerably in duration, depending on the discourse requirements of episodes in the story, and are frequently, but not always, separated by pauses.

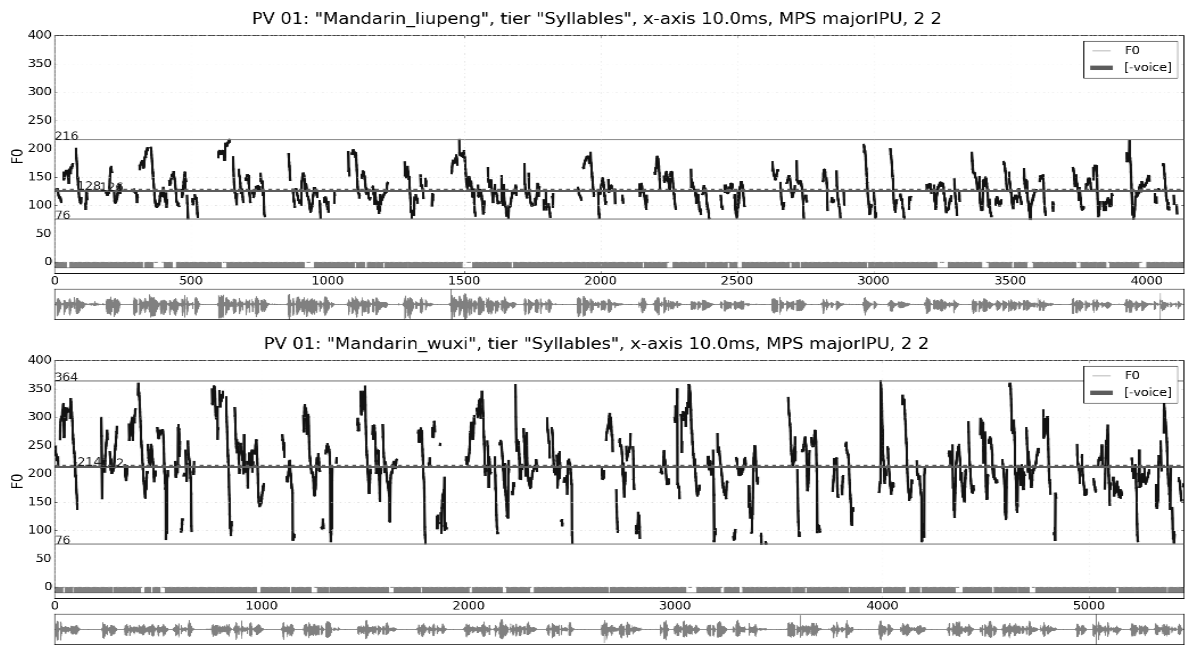
Figure 15: Male and female renderings of Aesop's fable *The North Wind and the Sun* in Mandarin.

Both speakers remain within fairly well-defined limits: below 216 Hz for the male voice and below 343 Hz for the female voice. The lowest value of 76 Hz in each case is an artefact due to the setting the minimal F0 for analysis at 70 Hz. In fact the lowest frequency, due to paratone-final creaky voice, is much lower than this, but the specific structure of creaky voice waveform is not visible in Figure 15 because of the long time scale. Careful inspection shows that there are potentially two F0 'bottom lines', though the higher level is not indicated explicitly: the higher level for the male voice is at about 100 Hz for the male speaker, and just below 150 Hz for the female speaker, typical levels for male and female speakers. The absolute lower level bottom line is for creaky voice.

The male range is quite typical of longer, narrative contexts. Paratones with their high initial F0 and final creaky voice cannot be captured by linguistic studies which restrict themselves to sentences or *faux* dialogues. The paratone flat hierarchy is quite well known to discourse analysts, however.

Another tutorial anecdote: The use of creaky voice at the paratone level of the discourse intonation hierarchy, is at the root of a phenomenon which has been commented on fairly extensively by the lifestyle media in recent years under the name 'vocal fry': the use of creaky voice by West and East Coast American female speakers to gain status in public by getting their voices to sound more like low male voices. This is a highly speculative interpretation by the media, of course, and not a view which can be confirmed without considerable empirical sociolinguistic and phonetic investigation. In fact it is easy to find amusing cases of critics who have unwittingly used vocal fry in their own critiques. In any case, the use of creaky voice by the Mandarin Chinese speakers, shown in a previous section, indicates that the creaky voice convention is certainly not restricted to young female adult Californians and Bostonians.

For formal and theoretical linguistics, an interesting property of this flat hierarchy, of two linear levels of paratones consisting of intonation units and of intonation units consisting of pitch accents and boundary tones, is that the pattern is recursive, but only in the simple 'cyclic finite automaton' sense of recursion, equivalent to loop iteration (Pierrehumbert 1980). Finite state or *linear recursion* only needs finite memory and linear processing complexity, unlike the more powerful *centre embedding* type of recursion, which requires polynomial complexity, typically of degree 3, and in principle unlimited memory (see Carlsson 2010 for further discussion of finite memory recursion in language). Whether prosodic patterns permit a more complex kind of recursion than the two-level linear loop recursion of these examples has been the subject of continuing debate in the literature for over fifty years.

## 6.2 English newsreading

In English newsreading monologues, very similar structures to the Mandarin story-telling monologues, which are of similar length, may be observed, as Figure 16 shows when compared holistically with Figure 15. The pitch patterns are organised into downward trending shorter F0 patterns, which are in turn merged into downward trending paratone F0 patterns, often separated by pauses, though the pauses are not as frequent or as long as in the Mandarin Chinese story-telling genre. The paratones are longer than in the Mandarin Chinese narrative examples, but the two-level flat hierarchy is similar.

Careful inspection of the frequencies between the bottom line and the zero line in Figure 16 (the beginning of extract A0101B from the Aix-Marsec Corpus, Auran et al. 2004) also suggests that the English female speaker also has both a 'creaky voice bottom line' and a 'regular bottom line', like the Mandarin Chinese speakers. This female English newsreader also tends to use creaky voice at the end of paratones. One very striking feature of this voice is the wide F0 range, with an extremely high paratone-initial F0. The highest paratone-initial F0 is 433 Hz and other paratone-initial peaks are around 400 Hz. Another striking feature is paratone-final creaky voice. Higher paratone-initial F0 and low paratone-final F0, especially creaky voice, thus appear to be a cross-language intonational convention, not limited to either tone languages or stress positional pitch accent languages, though not necessarily with the exact range of properties shown in these examples.

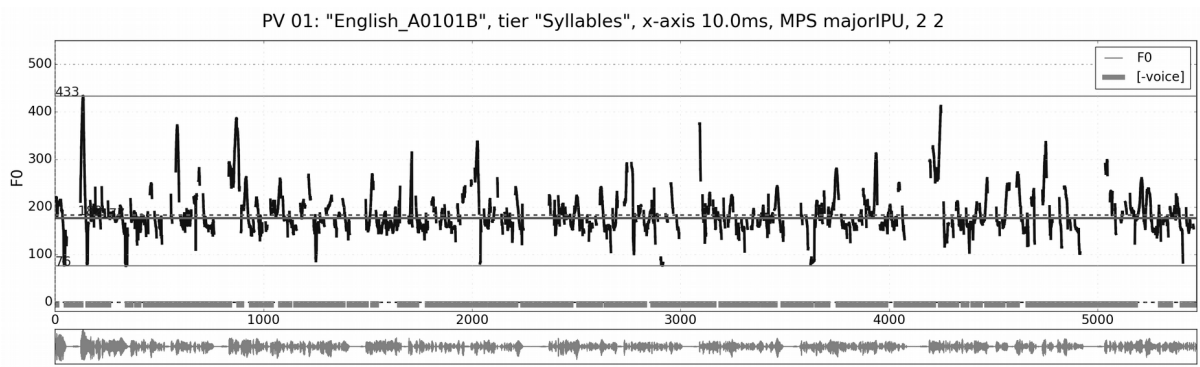

Figure 16: Female BBC newsreader, recording A0101B of the Aix-Marsec corpus of spoken English.

## 6.3 English interview: Question-Answer sequence

It is commonly assumed that questions in English end with a rising F0 contour and informative answers with a falling contour, with the exception of wh-questions. As with other melodic and rhythmic patterns, these assumptions reflect tendencies, rather than hard-and-fast rules, and may be overridden by more general semantic or pragmatic considerations.

The top and centre visualisations in Figure 17 provide a partial illustration of these general assumptions, in a formal radio BBC interview (extract J0104G from the Aix-Marsec corpus of BBC radio broadcasts from the 1980s, Auran et al. 2004). The assumptions are not fully borne out, however: the final contour of the answer is slightly rising. This is a case of overriding by a stronger semantic constraint: the second speaker continues by explaining his reply (not shown in Figure 17), and this small rise may be analysed as an addressee-oriented continuation marker, not as an interrogative.

The example illustrates not only a point about rising terminal contours in questions and continuation contours, but it also illustrates four further discourse properties of English intonation in formal styles of this type. These three properties have been noted in the older pedagogical literature (cf. the overview in Gibbon 1976) but not in more recent phonetic studies. Clearly, single examples such as those given here for tutorial purposes are not proof, or even indications of a tendency; however, they stand as prototypes representing hypotheses for further analysis. The further points are:

1. In the question, the pitch accents immediately before the final *rising* contour are L-accents, that is, they are F0 movements which first move downwards, then back up to the reference line.
2. It is not only the final contours of question and answer which rise and fall, respectively. The *polynomial global reference line* of the *question* is practically level, with a slight rise, as the superimposed model shows, mainly conditioned by the final rise. The global reference line of the *answer* is falling, also shown by the superimposed model, but with a much greater falling trend. The overall linear slope of the question contour is very small, 0.04, confirming a level pattern, with a standard deviation of 14.9. The overall linear slope of the answer contour is -0.27, a falling trend of much larger magnitude, with an expected larger standard deviation of 19.0.
3. The *overall height* of the *question* is on average higher than the overall height of the *answer*. Measurements of the raw pitch contour an F0 maximum of 205 Hz, mean of 151 Hz, median of 150 Hz and minimum of 115 Hz for the question, and 78 Hz, 114 Hz, 111 Hz and 84 Hz for the answer. The comparison is tentative, because further investigation is needed to show whether this particular feature is a general convergence strategy of the responder, or speaker-specific.

4. The question and answer are structurally linked by the overall *gestalt* of a rising-falling global contour which extends smoothly over both question and answer. The superimposed model displays as a smooth rising-falling curve connecting the question and the answer. More justification is needed for a gestalt claim of this kind, of course, but the pattern shown by the model is a useful heuristic pointer towards future study.

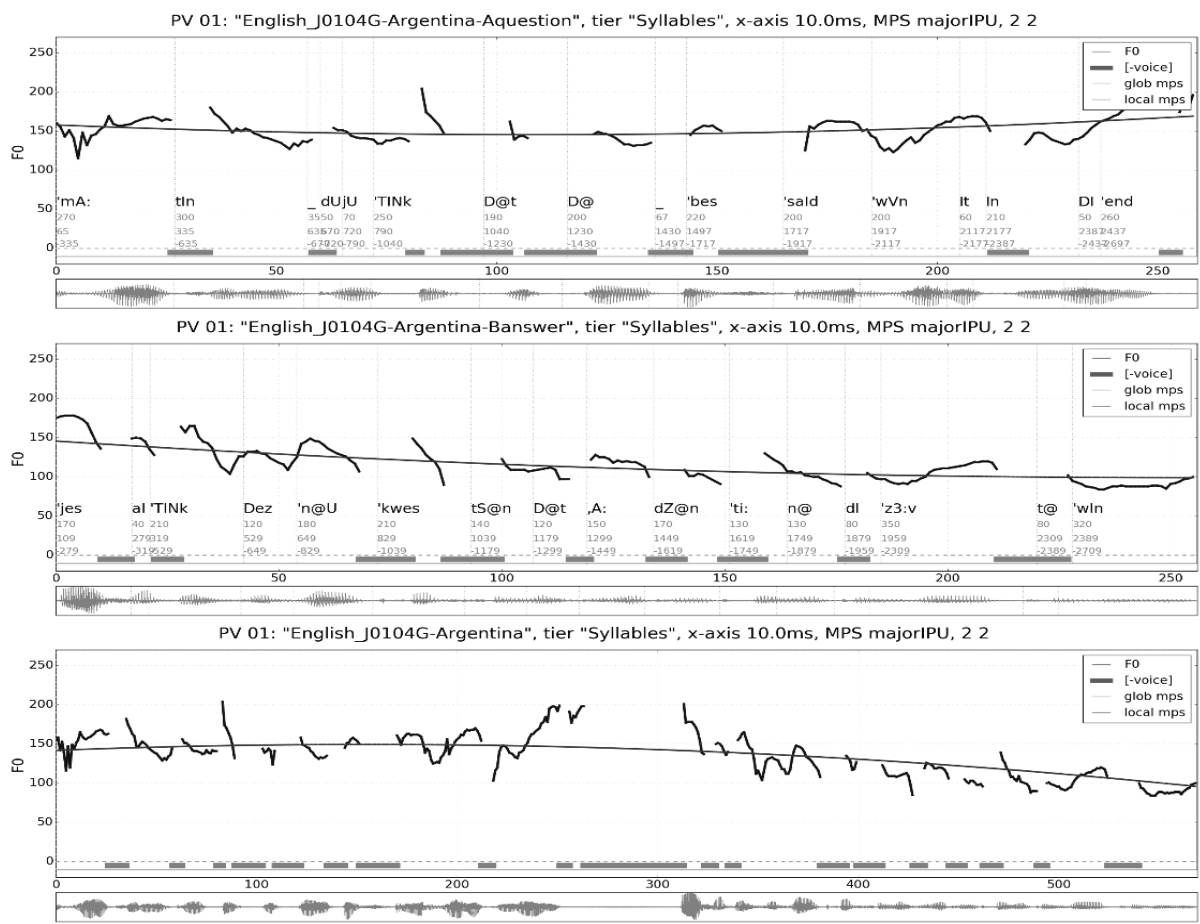

Figure 17: English question-answer interview sequence extracted from Aix-Marsec corpus, item J0104G (male voices).

A coherent functional interpretation of the four additional non-local properties is that they provide concomitant supporting markers for a questioning component which is nonfinal in the question-answer adjacency pair, and an answering sequence which is also nonfinal, but within the answer itself, with both non-final elements marked by final rising F0 patterns at different F0 heights.

### 6.4 When discourse starts and discourse stops: chanted contours ('call contours')

Chanted contours, or 'chromatic contours', which occur in approximately similar form in many otherwise unrelated languages, form a special case of monologue discourse intonation patterns, with their own functional, structural and phonetic characteristics. The most recent example of research on 'call contours' is the analysis of phonetic properties of Polish call contours by Arvaniti et al. (2016), who include a broad discussion of previous research.

 The contextual distribution of chanted contours is highly constrained to certain discourse contexts and is not usually concatenated with other intonation contours, except for humorous or ironic purposes. The term *chroma* was introduced (Gibbon 1976) to capture the musical quality of these contours. The mnemonic expression *call contour* which is often used does not describe the phonetic contour but one of its functions, and therefore begs the question of whether contours with calling function share similar forms or not: a *notional fallacy*.

 One well-known use of chanted contours is in children's taunts: *Cowardy cowardy custard, your face is made of mustard* (with many regional variants). In everyday discourse the chanted contours are typically used to start a discourse encounter by calling a potential interlocutor with a greeting such as "Hello" or "Good morning", as in the left-hand IPU (interpausal unit) in Figure 18, or by name. The speaker may not even know whether a potential interlocutor is present, as in the centre IPU in Figure 18, in which the chanted contour is used twice. These discourse-initiating functions are not the only uses of chanted contours. The chanted contours can also be used to terminate a discourse encounter, with farewells such as "By-ye!" or "Thank you" (for instance in a shop after a purchase), as in the right-hand IPU in Figure 18.

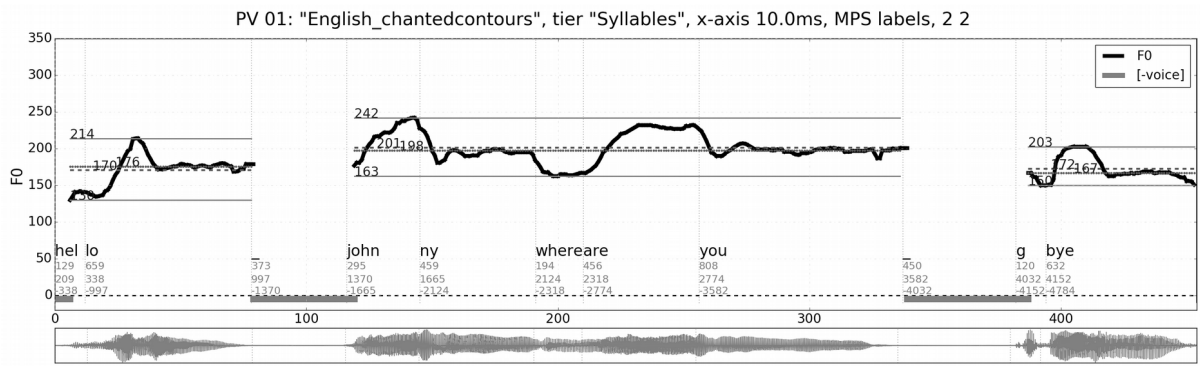

Figure 18: Three contexts for chanted contours in English (male voice).

The internal structure of the chanted contour has three components: a low-to-high unstressed initial section or *anacrusis*, a high and often level F0 section on the stressed syllable, $F0_1$, followed by a lower level F0 section until the end of the utterance, $F0_2$. These three components can be spread over several syllables, or crowded together on to one syllable, as in a call with a monosyllabic name: *Jo-ohn!* It is well-known that the phonetic properties of the F0 contour, apart from involving sustained level contours, include a fairly regular musical interval, stepping down by one *minor third*, spanning three semitones, between the second and third components of the contour. In English, when the chanted contour is repeated the repetition may occur with a rising musical interval of about a *sixth*, spanning nine semitones, i.e. three minor thirds.

*Table 2: F0 values and ratios for English chanted contour samples*

| Associated locution | $F0_1$ mean level | $F0_2$ mean level | $F0_1$:$F0_2$ actual ratio | Theoretical minor 3$^{rd}$ ratio | |
|---|---|---|---|---|---|
| | | | | just | tempered |
| hello | 212 | 177 | 1.198 | 1.2 | 1.189 |
| goodbye | 201 | 168 | 1.196 | | |
| Johnny | 240 | 196 | 1.224 | | |
| where are you | 230 | 197 | 1.168 | | |

The results for each of the chanted contours in Table 2 show slightly different F0 means for the examples displayed in Figure 18, in both the $F0_1$ the $F0_2$ components. However, the $F0_1$:$F0_2$ ratios of the means remain remarkably constant, and are very close to 1.2. The two slightly different theoretical ratios on the right of the table correspond to two ways of defining musical scales, which were already introduced in the discussion of F0 measurement and perception. The *just* scale is defined in terms of ratios between harmonics of the base note of the scale, while on the *tempered* scale the intervals have equal ratios, based on the ratio of 2:1 (the *octave* interval), i.e. the interval between the second harmonic and the base frequency. The tempered minor third is a quarter of an octave, and corresponds to an interval ratio of 1.189:1. Applying the ratio 1.189:1 four times yields the octave: $1.189 \times 1.189 \times 1.189 \times 1.189 = 1.189^4 = 2.0$. The tempered semitone, which was mentioned previously in connection with the logarithmic musical intervals which are relevant for modelling pitch perception, is one 12$^{th}$ of an octave, with a ratio of $2^{1/12}$:1, which approximates to 1.0595:1. There are 3 tempered semitones in a tempered minor third: $1.0595 \times 1.0595 \times 1.0595 = 1.0595^3 = 1.189$. Logarithmic ratios thus turn out not only to be a useful model of the human hearing scale, but also a phonetic feature at discourse rank.

The discourse rank context for the calling and greeting functions of chanted contours may be modelled as a discourse frame, in which the chanted contours play their role at the beginning and the end of the frame when there is a transition from noncommunication to communication or the reverse, either in calling and greeting to start communication, or in taking farewell in order to terminate communication. A functional explanation of the use of quasi-musical chanting may be that chants are typically used as a clearly audible signal in communication-at-a-distance, or *teleglossia*, in order to start or finish communication. The functions speech surrogate of whistling also includes the calling function with the same F0 pattern, though the absolute F0 is much higher, often 1000 Hz or more.

The discourse functions of the chanted contour can be partly specified in terms of the speech act theory of Searle, who defined a set of conditions for successful speech acts. The relevant condition for the chanted contour used as a call is the *normal input and output condition* (Searle 1969:57):

1. Normal input and output conditions obtain.

I use the terms "input" and "output" to cover the large and indefinite range of conditions under which any kind of serious and literal linguistic communication is possible. "Output" covers the conditions for intelligible speaking [...] and "input" covers the conditions of understanding. Together they include such things as that the speaker and hearer both know how to speak the language; both are conscious of what they are doing; they have no physical impediments to communication, [...] such as deafness, aphasia, or laryngitis; and they are not acting in a play or telling jokes, etc. It should be noted that this condition excludes both impediments to communication such as deafness and also parasitic forms of communication such as telling jokes or acting in a play.

The point is that in teleglossic situations such as calling, *Searle's normal input and output conditions do not obtain*. The contact has not yet been made, and the potential interlocutor may be too far away for normal communication, or indeed not present at all. The *lack of fulfilment* of the normal input and output condition is a necessary condition for use of the chanted contour in calls (Gibbon 1976). It is not a sufficient condition, of course, since there are also constraints on form, content and idiomaticity, and also since there are many other kinds of non-normal input and output condition.

German has an interesting generalisation of the discourse frame which does not occur in familiar varieties of English. When communication breaks down another kind of transition situation between communication and non-communication arises, and in this kind of transition the chanted contour may also be used. A personal anecdote may illustrate this. During my first class as a young lecturer in Germany I must have been speaking rather quietly, because someone at the back of the room shouted "Lauter!" (louder) with the chanted contour. My first reaction was that this was rather rude, until I realised, after more such experiences, that this use was a systematic convention and not a spontaneous impolite complaint. A discourse frame covering both English and German contexts is shown in Figure 19.

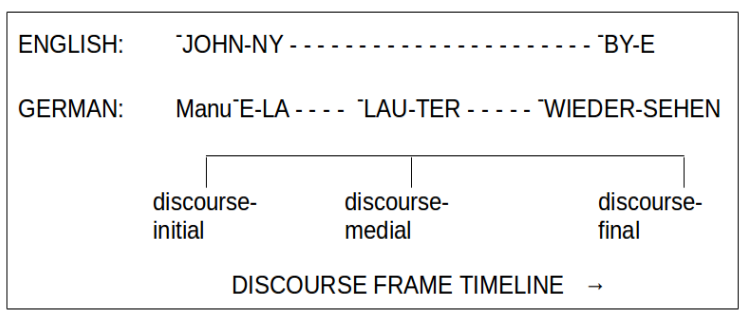

Figure 19: Discourse frame timeline as a partial model of the functionality of chanted intonation contours in English and German.

## 7 Discourse and speech timing: tones, stress, discourse rhythm

### 7.1 Tones, stress and timing in discourse contexts

There are many factors which affect speech timing in local syllable domains. One question which might be asked is whether the Mandarin Chinese tones tend to co-occur with different syllable durations. Because of the other factors involved it is not obvious that this will be the case, but it is a possibility. An analysis of the duration of tones in a monologue discourse context was conducted, based on the male voice which was illustrated in Figure 15 ( Gibbon and Yu 2015).

The result of the tone duration analysis is shown as a box-and-whisker plot in Figure 20. The tones are the independent variable on the horizontal *x*-axis and the tone durations are the dependent variable on the *y*-axis. The boxes enclose the middle 50% of the data points (the data points between the first and third quartile) and the middle line indicates the median duration (2nd quartile). The dot indicates the mean duration for that particular tone. The vertical lines to the left of the boxes show error boundaries, and are not important for present purposes. The vertical lines of black dots show individual measurements and their distribution along the interval duration axis.

The plot shows no significant difference in this data set between the four main tones (based on direct comparison of the boxes, which overlap very strongly). The exception is tone 5, the context-dependent neutral tone, which is known to be tendentially shorter than the other four tones. In order to establish whether the apparent minimal differences between tone 2, the rising tone, and the tones 1, 3 and 4 is significant a much larger data set needs to be examined.

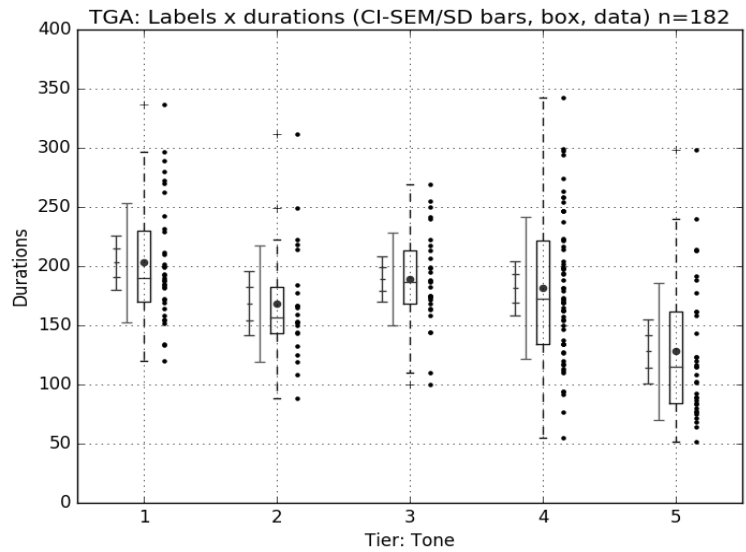

Figure 20: Mandarin Chinese syllable durations as a function of tone, from the CASS corpus (male voice).

The situation with syllable duration as a function of stress type in English for syllables with primary or secondary or no stress is quite different from the tone-duration relation in Mandarin Chinese. Traditionally, English syllables with primary stress are said to be longest, syllables which are unstressed to be shortest, and syllables with secondary stress to have durations between primary stressed syllables and unstressed syllables. This basic relation between syllable stress type and duration for genre F in the Aix-Marsec English corpus is tentatively confirmed, as Figure 21 illustrates, though the differences are not striking.

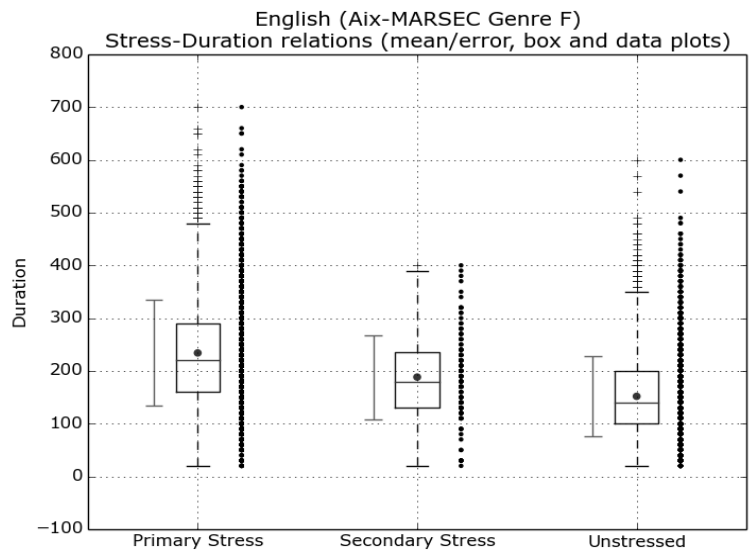

Figure 21: Syllable durations in magazine style radio reporting as functions of stress type (male voice, Aix-Marsec corpus, genre F).

The relations shown in Figure 20 and Figure 21 are tendencies for each of the stress positions, and not fixed rules. This is clearly shown by the vertical rows of black dots which visualise the broad dispersion of the durations of the individual syllables: there are unstressed syllables which are longer than most primary stressed syllables, and primary stressed syllables which are shorter than most unstressed syllables.

The broad distribution of durations for each category in Figure 20 and Figure 21 illustrates an important point about prosody in context, that the apparently clear lexical patterns found in citation contexts may be heavily influenced or even overridden by higher ranked semantic information structure and pragmatic emphasis, speech act sequences and emotionality. There are phonological factors such as the difference between phonemically long and short vowels and between syllable structures with simpler syllables (as in *ray*) compared with more complex syllables (as in *sprains*). There are phonetic factors such as whether a vowel borders on a voiced consonant or an unvoiced consonant, or whether there is utterance-final or paratone-final pre-pausal lengthening, or other general aesthetic and rhetorical principles of prosodic organisation such as strict rhythmic alternation. Analysis of duration over short domains like syllables must therefore be handled with great care.

## 7.2 The phonetic grounding of speech timing

English is often referred to as a *stress timed language* because of the role of stress positioning in creating an impression of rhythm when listening to the language. This is an oversimplification, however, as already noted, since there are many other factors which determine speech timing. In languages with a more regular and less varied syllable structure, such as Mandarin Chinese, the rhythmic effects are different, often being referred to as *syllable timing*, again an oversimplification.

Rhythms are repetitions of alternating values of some feature or features of speech (such as the intensity, duration or melody of syllables, words or phrases) at approximately equal time intervals, which play a role in the aesthetics and rhetoric of speech, and differ somewhat from one language or language variety to another under the influence of syllable, word, phrase, sentence, text and discourse structure. Rhythm has received a great deal of attention in recent years, particularly in in the study of Germanic languages such as English and German (Sudhoff et al. 2006; van de Vijver and Vogel 2015). It has become fashionable to be pessimistic about finding an empirical physical grounding for rhythm, because rhythm has remained something of a puzzle and it has proved to be very difficult to identify consistent physical properties of the rhythm of speech. The general view is gradually emerging that rhythm is an *epiphenomenon* which is more likely to be an organising principle in perception processing and cognitive integration than a property of speech production. In spite of the difficulty in finding physical evidence, intuitions about stress rhythm are nevertheless quite strong and have frequently been used in linguistically motivated searches for rhythm in word and sentence structure, both in citation contexts and in authentic discourse.

But it is still an interesting challenge to continue the search for a physical grounding for rhythm impressions by investigating aspects of discourse timing. Rough physical measures of timing which point to aspects of rhythm have been developed, even if they are not fully adequate, and will be discussed in the following sections. Regularity of timing is termed *isochrony*. The mechanical isochrony of a pendulum, a clock or a machine, with zero irregularity of rhythmic intervals is not a feature of speech. The articulatory organs are a very strange type of machine: soft, flexible and damp, and not at all suited to precise isochony. Speech production can be trained in this direction, of course, and in fact professional singers can come very close to strict isochrony.

The timing properties of speech have been investigated at different ranks from discourse rank to phoneme rank for very many years, including the timing of turn sequences in discourse (cf. Włodarczak 2014). There have been several studies of oscillator models in order to capture both the alternation property and the isochrony property of rhythm patterns as realistically as possible (Cummins and Port 1998 and the comparative study by Inden et al. 2012). The possibility of finding alternating rhythms at different ranks suggests that the right way to go is to model the production of rhythm as a set of highly variable oscillators which generate frequencies which are typical of different ranks, for example sequences of phrases at frequencies around one every 2 seconds (0.5 Hz), lexical words at about 1.5 per second in English (1.5 Hz), sequences of syllables at about 4 per second in English (4 Hz) and sequences of phonemes in English at about 10 per second (10 Hz). These numbers are informal illustrations of a possible general principle of timing in speech processing, not results of a specific empirical investigation.

## 7.3 Timing irregularity metrics

Several relatively recent studies of speech timing have developed measures for the isochrony property of rhythm without considering the alternation property, and are therefore by definition not adequate rhythm models since one necessary condition is missing (Gibbon 2003). The measures of speech timing developed in these studies are still useful pointers to differences in the regularity or irregularity of speech timing in different languages. In ignoring the rhythmic alternation property the measures have tended to concentrate on the relative duration of syllables or feet or of vocalic and consonantal intervals in speech, and on the global averages of such values. The global averages may evidently assign the same value to collections of duration values in any order, without regard to local alternation (Gibbon 2006). Several such measures have been proposed, and a selection will be discussed here.

One obvious measure of relative isochrony or its inverse, irregularity, is the *standard deviation* of durations of selected units relative to their mean duration: the smaller the standard deviation, the more regular the timing of the unit being measured, such as the syllable. But standard deviation is a global measure which pays no attention to orderings such as alternation. The strongest factor in determining low or high standard deviation, for example for syllable durations, is the complexity or simplicity of the syllable phonotactics of the language, independently of phonetic measurements.

A similar measure of relative isochrony, which has similar properties to standard deviation, is the *coefficient of variation*. The coefficient of variation is variant of standard deviation which is normalised by dividing the standard deviation by the mean (which therefore also ignores any alternation component) and multiplying by 100.

A closely related but slightly different measure of relative isochrony or irregularity, is the family of *Variability Indices* (*VI*), which has several quite similar predecessors (cf. Deterding 2001). The most popular *VI* is the *Pairwise Variability Index* (*PVI*), with two variants: the raw *rPVI*, typically used to measure durations of consonantal intervals, and the normalised *nPVI*, typically used to measure durations of vocalic intervals. The *nPVI* is normalised in order abstract away from the effect of speech rate changes on vowel length (cf. Grabe and Low 2002; Nolan and Asu 2009). The *rPVI* on the other hand has been used as a measure of consonantal variability, on the assumption that consonants are less subject to the influence of articulation rate than vowels. However, there is no particular reason not to use the *nPVI* for both. The *PVI* measures have been applied to many kinds of speech unit, often without questioning the validity of the *PVI* as a measurement of rhythm (Fuchs 2016).

The *VI* family measures are often referred to as *rhythm metrics*, but they share a major disadvantage with standard deviation: they do not measure rhythm but only relative isochrony or irregularity. Consequently, a more appropriate term would be *speech irregularity metrics*. Nevertheless, studies based on relative isochrony or irregularity scales clearly represent a significant step forward in the phonetic grounding of earlier phonologically motivated simple binary or ternary classifications such as 'mora timed', 'syllable timed' vs. 'stress timed' languages (cf. discussion by Dauer 1983), and have produced new insights into similarities and differences between languages in respect of speech timing.

Consequently, it is crucial to understand exactly how measures like the *VI* family and the *PVI* subfamily work, in order to be able to judge not only their contribution to the classification of languages, but also to the understanding of regularity or irregularity in speech timing. First, the formal structures of the metrics will be discussed, and in the following subsection they will be illustrated informally with specific cases (English, Mandarin Chinese and Farsi).

The simplest *PVI* version, *rPVI*, uses the mathematical operation of *differentiation* (more accurately: *differencing*) on sequences of speech interval durations in order to find the first derivative, i.e. the *rate of change* of the durations, the *acceleration* and *deceleration*, between one duration and the next. However, the *rPVI* does not use the true derivative, as negative duration differences between neighbours are converted into positive differences by taking their absolute value. The *rPVI* is the average of these absolute differences.

The *nPVI* is very similar, but with two differences. First, the absolute difference between each pair of durations is normalised by dividing it by the average of the two durations. This filters out the effect of speech rate variation. Second, the overall average is multiplied by 100 (as with the coefficient of variation) and rounded, to make it more readable as a whole number rather than a fraction. This multiplication does not actually make it a percentage, which has sometimes been claimed, as the scale is non-linear and asymptotic: $0 \leq i < 200$. Deterding (2001) suggested that normalisation of the *VI* by the overall average duration of the sequence, like with standard deviation, is more suitable for spontaneous speech than normalisation by the average of adjacent interval duration pairs, at the expense of filtering out speech rate variation.

*Table 3: Formulae for regularity measures SD, CoeffVar and nPVI.*

|  | Standard deviation | Coefficient of variation | nPVI |
|---|---|---|---|
| *Formula*: | $\sigma = \sqrt{\frac{1}{N} \sum_{i=1}^{N} (x_i - \mu)^2}$ | $c_v = \frac{\sigma}{\mu}$ | $nPVI = \frac{100}{n-1} \sum_{i=1}^{n-1} \frac{|d_i - d_{i+1}|}{d_i + d_{i+1}/2}$ |
| *Description*: | Square root of variance, the average squared difference between each measurement and the mean. | Standard deviation divided by mean. | 100 × average absolute duration differences between adjacent intervals, normalised by division by average duration of the interval pair. |

The formulae for standard deviation, coefficient of variation and *nPVI* are shown in Table 3. The *nPVI* formula has been formulated in an equivalent but slightly different way from the formula commonly found in the literature, in order to support comparison with the other two formulae. The main differences between the formulae are as follows:
1. where the squared difference is used for standard deviation, the absolute difference is used for the *nPVI*, with a similar result;
2. '*n*-1' is used in the *nPVI* because the number of pairwise differences between elements in the sequence is one less than the number of elements in the sequence;

3. where division of the standard deviation by the mean is used for normalisation in the coefficient of variation, division by the mean of the pair of neighbouring elements in the subtraction is used for normalisation in the *nPVI*;
4. multiplication by 100 in the *nPVI* is not really necessary, but is introduced in order to be able to use rounded whole numbers instead of fractional values below zero (it is sometimes maintained that the '100' converts the *nPVI* value to a percentage, but this is not strictly true as the maximum asymptotic value of the *nPVI* is 200, not 100).

Unfortunately, taking the absolute values of the duration differences in *rPVI* and *nPVI* removes all information about duration alternations which contribute to rhythm, leaving only information about relative isochrony (or its inverse, relative irregularity). In standard deviation, squaring of differences does the same. With all of these indices, an index of zero means perfectly mechanical rhythm. Higher values mean 'more and more irregular', but give less and less information about rhythm. A fundamental problem arising from the use of absolute values is that many different kinds of duration sequence may yield the same result, not only alternating rhythmic sequences. The *nPVI*, for example, may yield the sme result for mixtures of alternation with other geometrical series:

npvi(2,4,2,4,2,4, ...) = npvi(2,4,8,16,32,16,...) = npvi(32,16,8,16,8, 4...) = 66.66'

Another issue which the *VI* family irregularity measures faces is that *difference* between adjacent items is a *binary* operation, that is, either strictly *iambic* or strictly *trochaic*, with two units per rhythmic beat. But languages like English have many non-iambic and non-trochaic rhythms, for example *unary rhythm*, with one unit per beat, as in *Big John swam fast past Jane's boat* (*spondaic*, 'syllable timed'), various kinds of *ternary rhythm*, with three units per beat, such as *Jonathan Appleby idolised Wellington* (*dactylic*), *it's a shame that the girl couldn't come* (*anapaestic*) or *Amanda's collection of funny umbrellas was stolen* (*amphibrachic*), or even more complex rhythms with more units per beat, as in fast speech. Subtle domain differences of the rhythm operation like these are far beyond the scope of the irregularity metrics, and are a topic of further research in the *oscillator and entrainment* paradigm of rhythm research (Cummins and Port 1998; Inden et al. 2012).

## 7.4 Applying timing irregularity metrics

Summarising differences between the metrics, in the *VI* family duration differences are calculated between neighbouring units and the absolute value is used, while in standard deviation the differences are calculated between each unit and the mean and the difference is squared in order to obtain an absolute value. In both cases the average absolute value is calculated. In Deterding's *VI* each difference is normalised by the overall mean duration while in the *nPVI* the average of the adjacent values is used, and in the coefficient of variation the standard deviation is normalised by division with the mean. In both the coefficient of variation and the *nPVI* the average is multiplied by 100.

Similarities and differences between the quantitative timing irregularity measures standard deviation, coefficient of variation and *nPVI* are best demonstrated by application to specific cases. Applying the measures to the answer in the English question and answer sequence shown in Figure 17 (centre), typical values for English result:

*SD*(170,40,210,120,180,210,140,120,150,170,130,130,80,350,80,320) = 82

*nPVI*(170,40,210,120,180,210,140,120,150,170,130,130,80,350,80,320) = 60

*CoeffVar*(170,40,210,120,180,210,140,120,150,170,130,130,80,350,80,320) = 50

The *VI* family and the standard deviation measures yield quite similar results when used to compare languages and language varieties (Gibbon and Fernandes 2005). These values are fairly typical for English, a so-called 'stress-timed' language, and differ considerably from values for a so-called 'syllable-timed' language like Mandarin Chinese, with typical values of *SD*=59, *CoeffVar*=34, *nPVI*=40, respectively. Comparison of the ratios of these values for the ratio $dur_{English}:dur_{Mandarin}$ confirms that the measures are not too different: 1.7:1 (both standard deviation and coefficient of variation) and 1.5 (*nPVI*). Calculations were obtained by automatic annotation mining with the *TGA* online tool (Gibbon and Yu 2015). The lower value for *nPVI* reflects normalisation for variations in speech rate.

Comparing different languages is the classic application of the irregularity metrics, but application to varieties of one language may also be made, starting with the null hypothesis that languages are homogeneous in respect of speech timing strategies, and trying to falsify the null hypothesis. The metrics were applied informally to Farsi data, consisting of readings of a translation of *The North Wind and the Sun* (Marzban 2015), as with the Mandarin Chinese and English data in previous examples. A small pilot application was made, with recordings and annotation data for two male and three female Farsi speakers. The objective of the pilot application was to demonatrate the feasibility of the method for future use with larger data sets, rather than trying to achieve statistically informative results on the such a small data set. A range of descriptive statistics for for males and for females and for each metric are shown in Table 4. Clearly, the values for this small data set cannot be taken as serious results, but they show potential for further investigation. The most relevant results are shown in bold, and the ratios for the irregularity metrics discussed above are underlined.

Table 4 shows a small gender difference in the standard deviation, with a ratio of 1.13:1. This contrasts with the normalised irregularity metrics, the coefficient of variation and the *nPVI*, which show extremely small gender differences, with 1.05:1 for the coefficient of variation, and only 1.02:1 for the *nPVI*. So there may be interesting differences between the metrics, but maybe there are not so many differences between the genders. A tentative explanation for the low *nPVI* ratio may be that there is greater global variation in the speaking rate of males and females, and that this affects the speed-normalised *nPVI* to a lesser extent than the standard deviation, with an intermediate value for the coefficient of variation. This suggests that it may be interesting to compare the speed rates between speakers, and between males and females.

*Table 4: Irregularity metrics SD, CoeffVar and nPVI of syllable sequences for 2 male and 3 female Farsi speakers reading a translation of Aesop's fable "The North Wind and the Sun", most relevant parameters bolded.*

| Syllables | Male | | Female | | | Comparison | | |
| --- | --- | --- | --- | --- | --- | --- | --- | --- |
| | sp1 (M) | sp2 (M) | sp3 (F) | sp4 (F) | sp5 (F) | mean (M) | mean (F) | ratio M:F |
| **Median rate (s):** | **5.1** | **5.75** | **5.54** | **5.78** | **5.75** | **5.42** | **5.69** | **0.95:1** |
| Min duration (ms): | 82 | 64 | 66 | 58 | 47 | 73.0 | 57.0 | 1.28:1 |
| Max duration (ms): | 398 | 512 | 329 | 387 | 386 | 455.0 | 367.3 | 1.24:1 |
| Mean duration (ms): | 206.83 | 186.97 | 184.96 | 180.13 | 183.04 | 196.9 | 182.7 | 1.08:1 |
| **Median duration (ms):** | **196.0** | **174.0** | **180.5** | **173.0** | **174.0** | **185.0** | **175.8** | **1.05:1** |
| **SD duration (ms):** | **69.8** | **71.7** | **54.7** | **66.3** | **66.5** | **70.75** | **62.5** | <u>**1.13:1**</u> |
| **CoeffVar duration:** | **33.7** | **38.4** | **29.6** | **36.8** | **36.3** | **36.05** | **34.2** | <u>**1.05:1**</u> |
| **NPVI duration:** | **38** | **46** | **35** | **43** | **45** | **42.0** | **41.0** | <u>**1.02:**</u> |

The average of the male median speech rate is 5.42 syllables per second (median syllable duration: 185 ms) with quite a large difference between the two males. Speaking rate for two of the females were very close to each other, and also to the faster talking of the males, with median of 5.78, 5.75 and 5.75 syllables per second. The median is preferred, in order to minimise the effects of very long and very short syllables. The overall $dur_{Male}:dur_{Female}$ syllable rate ratio is 0.95:1, corresponding to an average median syllable duration for male and female speakers of 1.05:1. If this result can be believed, the female speakers read aloud faster than the males, with the same text. With such a small data set, this is more fiction than fact, of course.

The two informal applications to English and Mandarin Chinese, and to gender differences in Farsi, show that the irregularity ratios could be accompanied by analyses of speaking rate in order to achieve insightful results. Speaking rate analyses could take word rate and other domains into account, in order to gain insights into other measures as well as the consonantal, vocalic and syllabic timing (Nolan and Asu 2009) to which the measures are usually applied.

## 7.5 Speech timing: the irregularity of irregularity

When applied to durations of speech items such as syllables, the global dispersion metrics such as standard deviation and *nPVI* are measures of relative isochrony or timing irregularity alone, rather than of alternating rhythm. Since these irregularity metrics apply across the board, they give the impression that their values apply relatively uniformly to the entire recording and, in some discussions in the literature, to the entire language. The global values can be relatively consistent for a given data set, as shown by the relatively low variation shown by the Farsi pilot application, but this is not guaranteed.

This second null hypothesis of homogeneity can easily be tested by applying the metrics to subsequences of the recordings, for example in a moving window consisting of a certain number of syllables. The homogeneity null hypothesis is that the *nPVI* calculated for each window has an *nPVI* which is consistently very close to the *nPVI* of the other windows, in particular to the *nPVI* of the neighbouring windows.

The Farsi annotation data for two males and two of the females described in Table 4 were analysed by moving a window 5 syllables in length through the recording in steps of 1 syllable at a time. At each step, the *nPVI* was calculated for the 5 syllable window. In order to have a measure of similarity between the windows, the mean, standard deviation and coefficient of variation for the *nPVI* measurements in all the window positions were calculated, with the expectation that the standard deviation would be very small in relation to the mean if the speech timing were relatively homogeneous, but would approaching the mean if the speech timing were relatively inhomogeneous.

The results of the windowing procedure yield an averaged *nPVI* for the male speakers which corresponds exactly to the globally measured *nPVI* (38 and 46), as in Table 4. For the female speakers the results are minimally different, possibly due to rounding errors (43 and 44, but in Table 4 43 and 45). The coefficients of variation are 34.2, 27.8 for the males, and 31.7, 37.2 for the females. These coefficients of variation are very large, indicating a high degree of inhomogeneity of *nPVI* values for each speaker. These high coefficients of variation may be taken as falsifying the null hypothesis of homogeneity of *nPVI* throughout the utterance.

The results for the four speakers are visualised in Figure 22, which shows the progress of the moving window, plotting the *nPVI* value at each point. It is evident that the local *nPVI* values vary wildly, not only overall but also at very close ranges. The grey line from bottom left to top right in each graph represents the same *nPVI* values, but sorted in size from lowest to highest. The horizontal lines show the mean (centre) and the mean plus and minus the standard deviation.

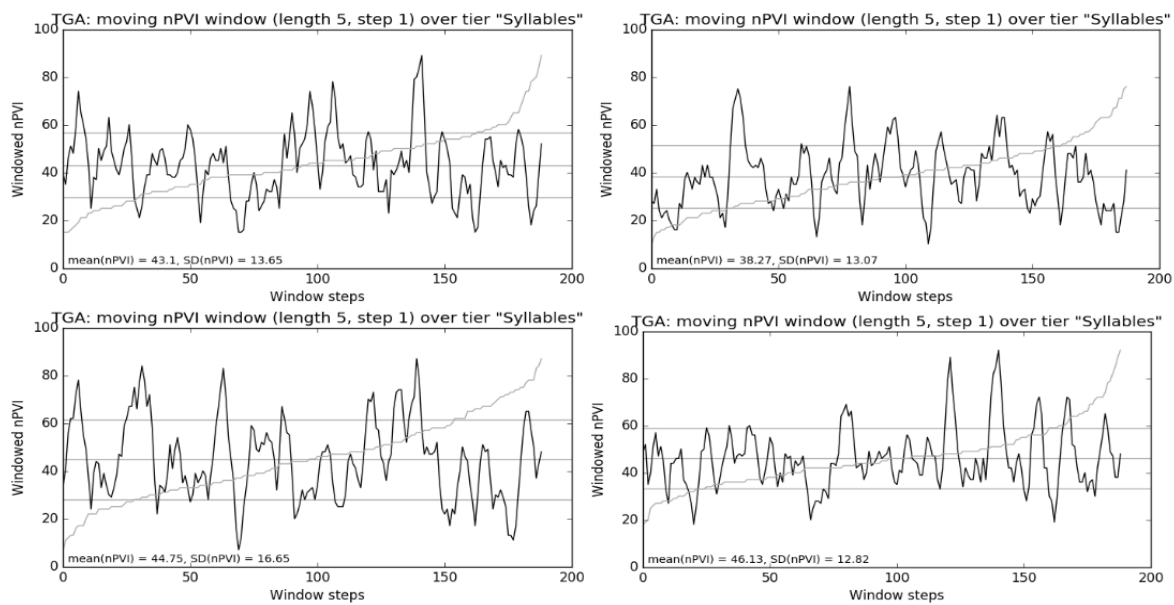

Figure 22: Moving *nPVI* window (length 5, step 1) for 2 female Farsi speakers (top and bottom left) and 2 male Farsi speakers (top and bottom right); gernerated with *TGA* software (Gibbon and Yu 2015).

Not only are the *nPVI* values heterogeneous through the utterance for each speaker, the patterns themselves are also, which would be expected, very different between speakers, even though the overall values were previously shown to be rather similar to each other. It is not unlikely that the speakers are using different rhetorical styles in reading, or have different degrees of practice at this kind of activity. The curves showing the sorted *nPVI* are very similar in each case, showing a fairly broad and even dispersion around the mean, and relatively few extremely low and extremely high values. Further analysis is needed in order to show how similar or different the timing irregularities are between the speakers, and to discover possible structural, semantic and pragmatic factors which determine the irregularities which were found.

The discovery of this inhomogeneity in speech timing is not necessarily a negative outcome for the irregularity metrics, but rather an illustration of why the study of speech rhythm is so elusive, with pointers to future research directions. Further investigation is likely to reveal relations between more and less regular passages and discourse strategies, as well as discourse control features such as disfluency. The fact that isochrony-irregularity metrics such as the *nPVI* fail to measure rhythm and are in themselves inhomogeneous is consequently not the end of the story for these metrics.

## 8  Closing remarks: summary, further reading and outlook

The preceding sections of this tutorial contribution on prosody are founded on a well-defined semiotic rank framework of the *Structure*, *Modality* and *Function* of prosodic signs, and on a structural *Rank Hierarchy* from discourse through text, utterance), sentence and phrase to the lexical prosody of morphemes and phonemes. The objectives of the tutorial are at once broad and restricted. The broad objective is to outline the domain of prosody: discussion, within the semiotic rank framework, of a somewhat eclectic selection of 'must know' topics in prosody from details of phonemic tone, accent in context, phrasal intonation and speech timing to the functionalities of prosody in discourse and prosodic differences between languages and language varieties. The narrow objective is methodological: to show quantitative phonetic measurements and visualisations in the acoustic phonetic domain, in the belief that not only does a picture convey more than a thousand words in the context of a tutorial, but that the precision and limitations of a heuristic measurement-and-visualisation-oriented approach offer a more fruitful long-term perspective than other approaches. Speculation about prosody in cognition is largely avoided, but theoretically interesting issues of prosodic universals figure by implication and brief discussion in each section.

Progress in the study of prosody has tended to be incremental rather than explosive, with new results coming in from many directions, from discourse analysis to phonetics and from computational linguistics to digital signal processing, with many cases of 're-discovery of the wheel' along the way. For further study there are many introductions, overviews, monographs, collections and conference proceedings which have appeared over a long period of time. The following brief list of books covers many aspects of the phonology, phonetics and the functionalities of tone and intonation: Couper-Kuhlen (1986), Crystal (1969), Féry (2017), Gibbon (1976), Gussenhoven (2004), Lehiste (1970), Ladd (2009), Pike (1945, 1948), Wells (2006) and Yip (2002). A wide variety of more phonetically oriented topics on both the rhythm side and the melody side of prosody is offered in the collections by Cutler and Ladd (1983), Gibbon and Richter (1984), Hirst and Di Cristo (1998) and Gibbon et al. (2012), mainly focussing on intonation, and by Fromkin (1978) and Jun (2005) on tone. A key study of timing is by Campbell (1992), and studies on rhythm are discussed in the collection by Sudhoff et al. (2006) and included in the collections by Gibbon et al. (2012) and van de Vijver and Vogel (2015). The other entries in the list of references for the present paper are also intended to help with further study.

Many important topics have not been covered in the tutorial. Prosody learning by children and adults of all ages, including prosody in second and third language learning, has been receiving increasing attention, and is leading to more insights into the universality of prosodic categories and patterns. A plausible prediction is that the current revival of interest in the evolution of language (as opposed to historical language change) will increasingly address the evolution of prosody, together with the evolution of music, since prosody and music are closely related basic components of human interaction, even more so than even the simplest combinations of words. One way of looking at the evolution of speech, perhaps the only empirically possibly way, is the traditional method of comparing human and animal communication. In the case of prosody this will be by meticulously modelling melodies and rhythms in animal communication, whether produced by phonation or by whistling, and comparing them with aspects of prosody such as those outlined in this tutorial. Fitch and Hauser (1995), for example, note that final rising and falling pitch appears to share some similar functions in nonhuman and human primates. There may be more involved than specific tones, however: the vocalisations of many mammals are varied but simple in structure. The calls, cries and hoots of some nonhuman primates can show complex iterations and combinations which appear to be linearly compositional, like human prosody, in that they merge recurring component patterns into recognisable 'flat hierarchies', somewhat like the two-level linear structure of paratone and intonation unit, and of intonation unit and pitch accent. But these are topics for another day.

## 9  Acknowledgements

I am very grateful to Prof. Qiuwu Ma and Dr. Jue Yu for the opportunity to present my ideas at the Tongji University Summer School, July 2016, on *Contemporary Phonetics and Phonology*, and to Prof. Qi Gong for hosting me as visiting professor at Jinan University, Guangzhou (Canton City). Sincere thanks go to staff and students both at the Tongji Summer School and at Jinan University, who have given me very perceptive and helpful feedback on the presentations and thereby helped to shape the present contribution. David Deterding, Judy Gilbert, Katarzyna Klessa, Stavros Skopeteas, Jolanta Bachan and Peng Li provided invaluable feedback on an earlier version of this paper. For uncountable discussions on African languages over several decades I would like to thank Prof. Firmin Ahoua, Abidjan, Côte d'Ivoire, Prof. Eno-Abasi Urua and Dr. Moses Ekpenyong, Uyo, Nigeria, Dr. Sophie Salffner, Dr. Nadine Grimm and Prof. Ulrike Gut, Münster. My particular gratitude goes to Dr. Jue Yu for our cooperative research on the prosody of Mandarin Chinese.

For support with gathering data and for discussion of the data, including instant responses on social media, I am heavily indebted to many colleagues, particularly Prof. Zakari Tchagbalé, Abidjan, Côte d'Ivoire, for Tem recordings based on his textbook (Tchagbalé 1984); to Dr. Sandrine Adouakou-Assané, formerly Bielefeld, for the Anyi/Agni data and for many discussions on tone and intonation in Agni/Anyi in the course of discussions of her dissertation (Adouakou 2005); to Dr. Marykim Haokip, Assam, for the Kuki-Thadou data, for detailed cooperation on Kuki-Thadou during my stay in New Delhi as Fellow of the Jawaharlal Nehru Institute of Advanced Study in 2009 and for discussions of her dissertation (Haokip 2015); to Peng Li, Guangzhou, for detailed clarification of the lexical data from Mandarin Chinese, and to Sheida Marzban, Tehran, for providing and discussing the Farsi data. Use of the Beijing CASS corpus for the Mandarin Chinese narratives (Li et al. 2000), and recordings of the Mandarin Chinese tones by anonymous speakers on the internet are gratefully acknowledged.

The prototype *PV* and *TGA* web apps used in this tutorial are currently located at the following temporary addresses:

```
wwwhomes.uni-bielefeld.de/gibbon/PV/    wwwhomes.uni-bielefeld.de/gibbon/TGA/
```

# 10 References

Adouakou, Sandrine. 2005. Tons et intonation dans la langue agni indénié. Dr.Phil. dissertation, Bielefeld University.
Arvaniti, Amalia, Marzena Żygis and Marek Jaskułac. 2016. The phonetics and phonology of Polish calling melodies. *Phonetica* 73.
Beckman, Mary E., Julia Hirschberg and Stephanie Shattuck-Hufnagel. 2005. The original ToBI system and the evolution of the ToBI framework. In: Jun, Sun-Ah., ed. 2005. *Prosodic Typology: The Phonology of Intonation and Phrasing*. New York: Oxford University Press.
Boersma, Paul 2001. Praat, a system for doing phonetics by computer. *Glot International*. 5:9/10, 341-345.
Bolinger, Dwight L. 1951. Intonation – levels versus configurations. *Word* 7.
Brazil, David, Michael Coulthard and Catherine Johns. 1980. *Discourse Intonation and Language Teaching*. London: Longman.
Carlsson, Fred. 2010. Syntactic recursion and iteration. In: Harry van der Hulst, ed., *Recursion and Human Language*. Berlin/New York: Mouton de Gruyter, 2010.
Couper-Kuhlen, Elizabeth. 1986. *Introduction to English Prosody*. London: Edward Arnold.
Crystal, David. 1969. *Prosodic Systems and Intonation in English*. Cambridge University Press.
Cummins, Fred and Robert Port. 1998. Rhythmic constraints on stress timing in English. *Journal of Phonetics* 24:145-171.
Cutler, Ann and D. Robert Ladd. 1983. *Prosody: Models and Measurements*. Berlin etc.: Springer-Verlag.
Dauer, Rebecca M. 1983. Stress timing and syllable-timing reanalyzed, Journal of Phonetics, 11, 51}62.
Deterding, David. 2001. The measurement of rhythm: a comparison of Singapore and British English. *Journal of Phonetics* 29.
Dutoit, Thierry. 1997. *An Introduction to Text-To-Speech Synthesis*. Dordrecht: Kluwer Academic Publishers.
Féry, Caroline. 2017. *Intonation and Prosodic Structure*. Cambridge: Cambridge University Press.
Fitch, W. Tecumseh and Marc. D. Hauser. 1995. Vocal production in nonhuman primates: Acoustics, physiology, and functional constraints on "honest" advertisement. *American Journal of Primatology* 37 (3): 191–219.
Fromkin, Victoria A., ed. 1978. Tone: a linguistic survey. New York: Academic Press.
Fuchs, Robert. 2016. *Speech Rhythm in Varieties of English. Evidence from Educated Indian English and British English*. Berlin: Springer.
Gibbon, Dafydd. 1976. *Perspectives of Intonation Analysis*. Bern, Lang.
Gibbon, Dafydd. 1987. Finite state processing of tone languages. P*roc. European ACL*, Copenhagen.
Gibbon, Dafydd 1995. Empirical and semiotic foundations of prosodic analysis. In: Uta Quasthoff, ed. (1995), *Aspects of Oral Communication, Research in Text Theory*. Berlin: de Gruyter.
Gibbon, Dafydd. 2001. Finite state prosodic analysis of African corpus resources. *Proc. Eurospeech 2001*, Aalborg, Denmark, I: 83-86.
Gibbon, Dafydd. 2003. Computational modelling of rhythm as alternation, iteration and hierarchy. *Proceedings of the International Congress of Phonetic Sciences*, Barcelona, August 2003, III: 2489-2492.
Gibbon, Dafydd. 2006. Time Types and Time Trees: Prosodic Mining and Alignment of Temporally Annotated Data. In: Sudhoff, Stefan, Denisa Lenertova, Roland Meyer, Sandra Pappert, Petra Augurzky, Ina Mleinek, Nicole Richter and Johannes Schließer, eds. (2006). *Methods in Empirical Prosody Research*. Berlin: Walter de Gruyter, 281-209.
Gibbon, Dafydd and Helmut Richter, eds. 1984. *Intonation, Accent and Rhythm. Studies in Discourse Phonology*. Berlin, de Gruyter.
Gibbon, Dafydd and Flaviane Romani Fernandes. 2005. Annotation-Mining for Rhythm Model Comparison in Brazilian Portuguese. *Proc. Interspeech* 2005. 3289-3292.
Gibbon, Dafydd, Pramod Pandey, Mary Kim Haokip and Jolanta Bachan. 2009. Prosodic issues in synthesising Thadou, a Tibeto-Burman tone language. *Proc. Interspeech 2009*. Brighton, UK.
Gibbon, Dafydd, Daniel Hirst and Nick Campbell, eds. 2012. *Rhythm, Melody and Harmony in Speech. Studies in Honour of Wiktor Jassem*. Special Edition of *Speech and Language Technology* 14/15. Poznań: Polish Phonetics Society.
Gibbon, Dafydd and Jue Yu. 2015. Time Group Analyzer: Methodology And Implementation. *The Phonetician*, 9-34.
Goldsmith, John. 1990. *Autosegmental and metrical phonology*. Oxford: Basil Blackwell.
Grabe, Esther, & Low, Ee Ling. 2002. Durational variability in speech and the rhythm class hypothesis. *Laboratory Phonology* 7: 515-546.
Gussenhoven, C. 2004. *The Phonology of Tone and Intonation*. Cambridge University Press.
Halliday, Mark A. K. 1967. *Intonation and Grammar in British English*. The Hague: Mouton.
Haokip, MaryKim. 2015. *Grammar of Thadou-Kuki: a descriptive study*. Ph.D. thesis, Jawaharlal Nehru University, New Delhi.
Hess, Wolfgang J. 1983. *Pitch Determination of Speech Signals - Algorithms and Devices*. Berlin, Heidelberg: Springer, Berlin, Heidelberg.
Hirst, Daniel J. and Albert Di Cristo, eds. 1998. *Intonation Systems. A survey of Twenty Languages*. Cambridge: Cambridge University Press.
Hirst, Daniel, Albert Di Cristo & Robert Espesser 2000. Levels of representation and levels of analysis for intonation. In: Merle Horne, ed. *Prosody: Theory and Experiment*. Kluwer Academic Publishers, Dordrecht.
Hirose, Keikichi, Hiroya Fujisaki and Mikio Yamaguchi. 1984. Synthesis by rule of voice fundamental frequency contours of spoken Japanese from linguistic information. *IEEE, ICASSP 84* Vol. 9.
Hyman, Larry M. 2009. How (not) to do phonological typology: the case of pitch-accent. *Language Sciences* 31:213–238.
Hyman, Larry M. 2010. Kuki-Thaadow: an African tone system in Southeast Asia. In Franck Floricic, ed. *Essais de typologie et de linguistique générale. Mélanges offerts à Denis Creissels*. Lyon: ENS Éditions. 31-51.
Inden, Benjamin, Zofia Malisz, PetraWagner, IpkeWachsmuth. 2012. Rapid entrainment to spontaneous speech: A comparison of oscillator models. Proc. *CogSci*.
Jassem, Wiktor and Dafydd Gibbon. 1989. Re-defining English stress. *Journal of the International Phonetic Association*. 10, 1980:2-16.
Jun, Sun-Ah., ed. 2005. *Prosodic Typology: The Phonology of Intonation and Phrasing*. New York: Oxford University Press.
Kager, René (1999). *Optimality Theory*. Cambridge: Cambridge University Press.
Klessa, Katarzyna. 2016. *Annotation Pro. Enhancing analyses of linguistic and paralinguistic features in speech*, Wydział Neofilologii UAM, Poznań.
Ladd, D. Robert. 2009. *Intonational Phonology*. 2nd Edition (1[st] edn. 1996). Cambridge: Cambridge University Press.
Lehiste, Ilse. 1970. *Suprasegmentals*. Cambridge, Mass.: MIT Press.
Lehiste Ilse. 1975. The Phonetic Structure of Paragraphs. In: Antonie Cohen and Sibout G. Nooteboom, eds., *Structure and Process in Speech Perception. Communication and Cybernetics*, vol 11. Berlin, Heidelberg: Springer.
Li, Aijun, Fang Zheng, William Byrne, Pascale Fung, Terri Kamm, Liu Yi, Song Zhanjiang, Umar Ruhi, Veera Venkataramani, and Xiaoxia Chen. 2000. CASS: a phonetically transcribed corpus of Mandarin spontaneous speech. Proc. ICSLP 2000 / Interspeech 2000, Beijing, China.
Li, Peng. 2017. The meanings of 'ma'. Jinan University, Guangzhou. Personal communication.
Marzban, Sheida. 2015. Farsi corpus "The North Wind and the Sun". Arak: Arak University.
Mehler, Alexander, Laurent Romary and Dafydd Gibbon, eds. 2012. *Handbook of Technical Communication*. Berlin: Walter de Gruyter.

Mertens, Piet. 2004. The Prosogram : Semi-Automatic Transcription of Prosody based on a Tonal Perception Model. In B. Bel & I. Marlien, eds.. *Proceedings of Speech Prosody* 2004, Nara (Japan).
Nolan, Francis and Eva Lina Asu. 2009. The pairwise variability index and coexisting rhythms in language. *Phonetica*, 66(1-2), 64-77.
O'Connor, Joseph D. and Gordon F. Arnold 1961. *The Intonation of Colloquial English*. (2$^{nd}$ edition.) London: Longmans.
Palmer, Frank R. 1969. *Prosodic Analysis*. Oxford: Oxford University Press.
Palmer, Frank F., ed. . 1968. *Selected Papers of J. R. Firth, 1952-59*. London: Longmans.
Pierrehumbert, Janet B. 1980. *The Phonology and Phonetics of English Intonation*. Ph.D. Thesis, MIT. (1987: Bloomington: Indiana Linguistics Club.)
Pike, Kenneth L. 1945. *The Intonation of American English.* Ann Arbor: University of Michigan Press.
Pike, Kenneth L. 1948. *Tone Languages: A Technique for Determining the Number and Type of Pitch Contrasts In a Language, with Studies in Tonemic Substitution and Fusion*. Ann Arbor: The University of Michigan Press.
Poser, William J. 1984. *The phonetics and phonology of tone and intonation in Japanese*. Ph.D. thesis, MIT.
Poser, William J. 2015. Using SOX. http://billposer.org/Linguistics/Computation/SoxTutorial.html (accessed 2017-03-29).
Sagisaka, Yoshinori, Nick Campbell, Nick, Norio Higuchi, eds. 1997. *Computing Prosody. Computational Models for Processing Spontaneous Speech*. Berlin: Springer Verlag.
Searle, John R. 1969. *Speech Acts*. Cambridge: Cambridge University Press.
Selkirk, Elisabeth. 1984. *Phonology and syntax: the relationship between sound and structure*. Boston, MA: The MIT Press.
Sudhoff, Stefan, Denisa Lenertova, Roland Meyer, Sandra Pappert, Petra Augurzky, Ina Mleinek, Nicole Richter and Johannes Schließer, eds. 2006. *Methods in Empirical Prosody Research*. Berlin: Walter de Gruyter.
Talkin, David. 1995. A Robust Algorithm for Pitch Tracking (RAPT). In: W. B. Kleijn and K. K. Palatal, eds. *Speech Coding and Synthesis*, 497-518, Elsevier Science B.V.
Tatham, Mark and Katherine Morton. 2011. *A guide to Speech Production and Reception*. Edinburgh: Edinburgh University Press.
Taylor, Paul. 2000. Analysis and synthesis of intonation using the Tilt model. *Journal of the Acoustical Society of America*. 107(3):1697-1714.
Tchagbalé, Zakari, ed. 1984. *T. D. de Linguistique: Exercices et corrigés. No. 103*. Abidjan: Université Nationale de Côte d'Ivoire, Institut de Linguistique Appliquée.
't Hart, Johan, René Collier and Antonie Cohen. 1990. *A Perceptual Study of Intonation*. Cambridge: Cambridge University Press.
van de Vijver, Ruben and Ralf Vogel, eds. 2015. *Rhythm in Cognition and Grammar. A Germanic Perspective*. Berlin: De Gruyter.
Wells, John C. 2006. *English Intonation: An Introduction*. Cambridge University Press.
Wichmann, Anne. 2000. *Intonation in Text and Discourse: Beginnings, Middles and Ends*. London: Longman.
Włodarczak, Marcin. 2014. T*emporal entrainment in overlapping speech*. Dr. Phil. Dissertation, Bielefeld University.
Xu, Yi. 2011. Post-focus compression: cross-linguistic distribution and historical origin. *Proc. ICPhS XVII Special Session*. Hong Kong.
Yip, Moira. 2002. *Tone*. Cambridge, UK: Cambridge Univ. Press.